\documentclass[5p]{elsarticle}

\usepackage{lineno,hyperref}
\usepackage{subcaption} 
\usepackage{bm} 
\usepackage{amssymb} 
\usepackage{mathtools}  
\usepackage{gensymb}    
\usepackage{commath} 
\usepackage[nameinlink,capitalize]{cleveref}   
\usepackage{tikz}   
\usepackage{booktabs,tabularx}   
\usepackage{xltabular} 
\usepackage{pdflscape} 
\usepackage{nicefrac}   
\usepackage{dblfloatfix}    
\usepackage{layouts} 

\emergencystretch=\maxdimen
\DeclareMathAlphabet\mathbfcal{OMS}{cmsy}{b}{n} 
\DeclareRobustCommand{\rchi}{{\mathpalette\irchi\relax}}
\newcommand{\irchi}[2]{\raisebox{\depth}{$#1\chi$}} 

\newcommand{\cross}[1][1pt]{\ooalign{   
  \rule[1ex]{1ex}{#1}\cr
  \hss\rule{#1}{.7em}\hss\cr}}

\DeclareMathOperator*{\argmin}{\text{\itshape argmin}}    

\newcommand*\circled[1]{\tikz[baseline=(char.base)]{
            \node[shape=circle,draw,inner sep=1pt] (char) {\footnotesize #1};}}

\newcolumntype{V}[1]{>{\raggedright\let\newline\\\arraybackslash\hspace{0pt}}m{#1}}

\newcolumntype{L}{>{\raggedright\let\newline\\\arraybackslash}X}

\hypersetup{%
  colorlinks,
  allcolors=blue
}

\modulolinenumbers[5]

\journal{XXXXXX}

\begin{document}





\begin{frontmatter}
\title{Sleep Posture One-Shot Learning Framework Using Kinematic Data Augmentation: In-Silico and In-Vivo Case Studies\tnoteref{mytitlenote}}



\author[1]{Omar Elnaggar\corref{cor1}}
\ead{Omar.Elnaggar@liverpool.ac.uk}
\author[2]{Frans Coenen}
\author[3]{Andrew Hopkinson}
\author[4,5]{Lyndon Mason}
\author[1]{Paolo Paoletti}

\cortext[cor1]{Corresponding author}

\address[1]{School of Engineering, University of Liverpool, Liverpool L69 3GH, United Kingdom}
\address[2]{School of Electrical Engineering, Electronics and Computer Science, University of Liverpool, Liverpool L69 3BX, United Kingdom}
\address[3]{School of Psychology,University of Liverpool, Liverpool L69 7ZA, United Kingdom}
\address[4]{School of Medicine, University of Liverpool, Liverpool L69 3GE, United Kingdom}
\address[5]{Department of Trauma and Orthopaedics, Liverpool University Hospitals NHS Foundation Trust, Liverpool L9 7AL, United Kingdom}





\begin{abstract}
Sleep posture is linked to several health conditions such as nocturnal cramps and more serious musculoskeletal issues.
However, in-clinic sleep assessments are often limited to vital signs (e.g. brain waves). Wearable sensors with embedded inertial measurement units have been used for sleep posture classification; nonetheless, previous works consider only few (commonly four) postures, which are inadequate for advanced clinical assessments.
Moreover, posture learning algorithms typically require  longitudinal data collection to function reliably, and often operate on raw inertial sensor readings unfamiliar to clinicians.
This paper proposes a new framework for sleep posture classification based on a minimal set of joint angle measurements. The proposed framework is validated on a rich set of twelve postures in two experimental pipelines: computer animation to obtain synthetic postural data, and human participant pilot study using custom-made miniature wearable sensors. Through fusing raw geo-inertial sensor measurements to compute a filtered estimate of relative segment orientations across the wrist and ankle joints,  the body posture can be characterised in a way comprehensible to medical experts. The proposed sleep posture learning framework offers plug-and-play posture classification by capitalising on a novel kinematic data augmentation method that requires only one training example per posture.
Additionally, a new metric together with data visualisations are employed to extract meaningful insights from the postures dataset, demonstrate the added value of the data augmentation method, and explain the classification performance.
The proposed framework attained promising overall accuracy as high as $100\%$ on synthetic data and $92.7\%$ on real data, on par with state of the art data-hungry algorithms available in the literature.



\end{abstract}


\end{frontmatter}


%
%

\section{Introduction}  \label{sec:introduction}
A recent comprehensive epidemiological study revealed that nearly $22\%$ of the global population suffer from musculoskeletal disorders with most cases being in high-income countries \cite{Cieza2020a}. For example, in the United Kingdom, musculoskeletal conditions affect 1 in every 4 adults. One-third of medical consultations \cite{DepHealth2006} and over $25\%$ of all surgical interventions \cite{Clark2014} are consequent to musculoskeletal conditions. Another study projects these conditions will rise more rapidly in low- and middle-income countries \cite{Hartvigsen2018}.

The study of human posture allows for understanding the musculoskeletal system and opens the door for supporting the musculoskeletal health and well-being over the whole lifespan. Over recent years, human sleep behaviour studies have gained more traction among the research community \cite{Abdel-Basset2020}. Traditionally, sleep had been considered as a natural mechanism to recover from exhaustion of daily activities, but recent sleep studies gave contradictory observations. In fact, it was found that certain sleep behaviours could bring about health complications, such as {\itshape pressure ulcers} \cite{Paquay2008}, or uncover underlying disorders \cite{Ibanez2018}, including {\itshape restless leg syndrome} and {\itshape periodic leg movements}. Interestingly, some studies linked musculoskeletal morbidity to postural cues, for example, the supine position has been correlated to apnoea more strongly compared to lateral positions \cite{Pinna2015}. Another evidence shows that prolonged joint immobilisation could lead to muscular contractions \cite{Akeson1987} which could potentially develop into chronic pain episodes. Moreover, muscle cramps and painful spasms can also occur during wake or sleep states due to sustained abnormal body postures, lack of exercise or pregnancy \cite{Parisi2003}.

Motivated by the evidence above, clinicians and biomedical engineers are keen to investigate whether a significant statistical link exists between the development of musculoskeletal diseases and specific sleep postures. To this end, body sleep postures need to be monitored by means of {\itshape motion capture} technologies, which are generally categorised into {\itshape optical} and {\itshape non-optical} techniques, both of which have been employed to monitor the sleep body postures.

Optical methods are categorised according to whether they involve the use of on-body retroreflective markers or not: marker-based versus markerless techniques. The first category tends to be impractical for sleep analysis due to cost of the equipment involved, controlled lab setting requirements, and marker occlusions. Markerless motion capture capitalises on recent advances in computer vision and deep neural architectures to regress over the body-surface coordinates given a set of image pixels \cite{Akbarian2019, Li2018d, Mohammadi2018}. Markerless techniques also struggle with occlusions due to body covering and are often criticised over privacy failing, thus limiting their adoption.

Non-optical motion capture methods in sleep-related applications comprise two main categories; {\itshape bed-embodied sensors} and {\itshape wearable sensors}. Within the former category, force sensitive resistor grids embedded into mattresses \cite{Kim2019} and load cells attached to bed frame supports \cite{Alaziz2020}, are by far the most common techniques. However, bed-embodied sensors only provide measurements of the body weight distribution, which consequently require an indirect pose inference framework that is not guaranteed to be entirely reliable. Wearable inertial sensing offers a solution with low intrusiveness, does not require optical line of sight and guarantees privacy, addressing the aforementioned limitations of both optical and bed-embodied non-optical techniques. Moreover, processing of low-dimensional timeseries from on-body sensors is generally of low computational cost. Therefore, it is overall more suited for sleep monitoring applications.


There are a number of open research questions that hinder the large-scale deployment of wearable inertial sensors for tracking sleep postures. The challenges are primarily with the sensing and intelligent perception aspects of these systems. Measurement errors \cite{Qureshi2017}, sensor misalignment with respect to body segments \cite{Fan2022}, and {\itshape soft tissue artefact} \cite{Leardini2005} are amongst the most prominent sensing errors. Speaking of intelligent perception, we herein focus on three main challenges. First, wearable inertial sleep trackers have so far been exploited mostly for standard posture sensing (supine, prone and lateral positions) which has little to offer clinicians studying posture-dependent musculoskeletal pathologies, such as leg and calf cramps. Second, current works typically employ {\itshape machine learning} (ML) models that directly operate on raw sensor data - an incomprehensible black-box framework to clinicians who have an outsider perspective on artificial intelligence. Third, for these models to function reliably, extended data collection and expensive manual labelling are often prerequisites.

\begin{figure*}[p]
    \centering
    \includegraphics[height=0.97\textheight]{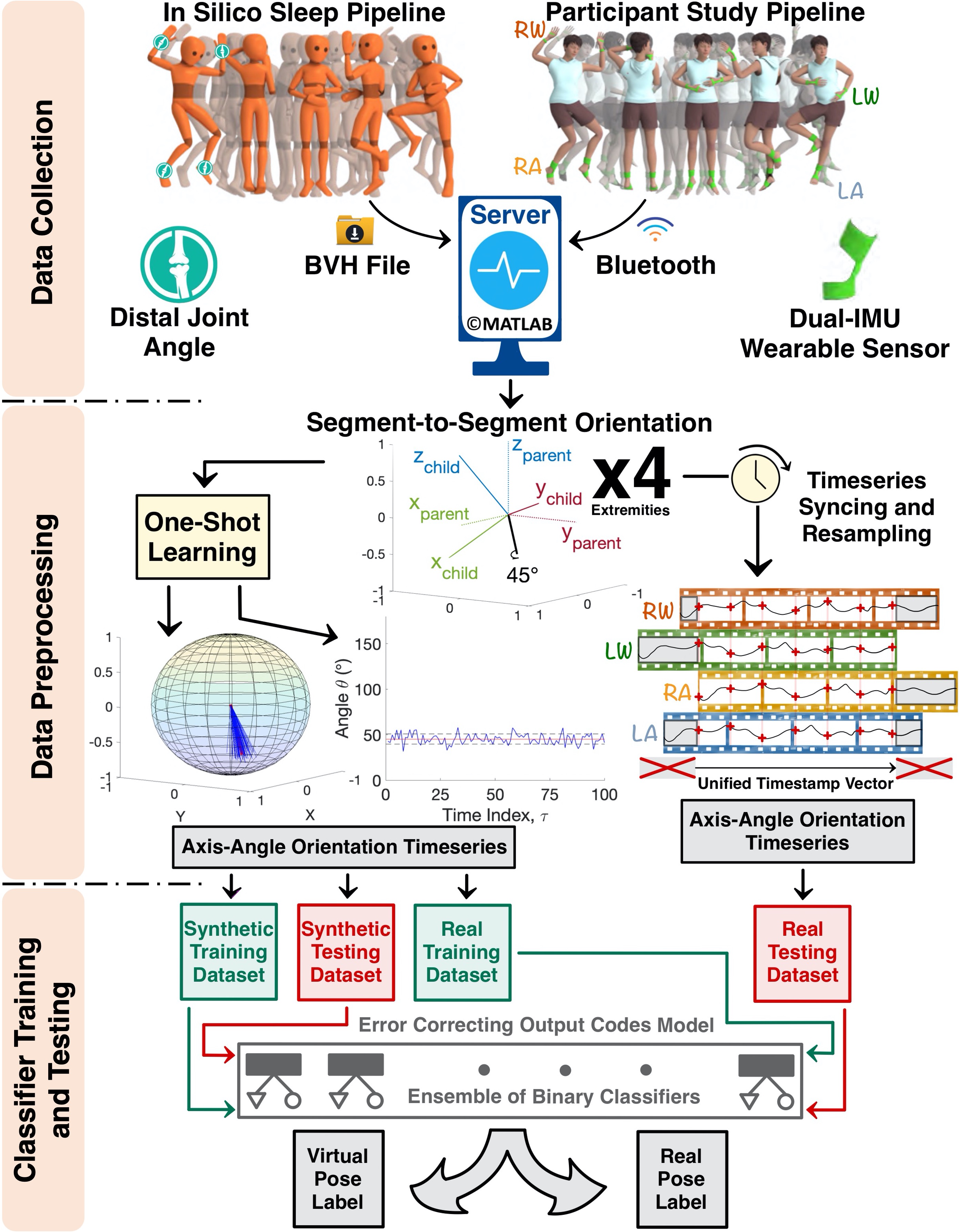}
    \caption{Schematic of the proposed sleep posture classification framework.}
    \label{fig:systemfigure}
\end{figure*}

This paper proposes a human sleep posture learning framework (illustrated in \cref{fig:systemfigure} and detailed in \cref{sec:methods}) to overcome the aforementioned challenges. The framework capitalises on data augmentation to facilitate sleep posture modelling from a single postural observation (hereafter ``shot''). The experimental pipelines have been developed and validated both {\itshape in silico} and on real world data.
The main contributions of the presented work can be summarised as follows:

\begin{itemize}
\item Our approach is the first study directed at wearable-based classification of twelve sleep postures, whereas previous work had been mostly limited to four ``standard'' postures. The twelve postures include a much wider range of postures common in sleep, thus making the proposed framework better suited for clinical use.
\item To the best knowledge of the authors, we are the first to use {\itshape inertial sensor fusion} in sleep postural analysis. Unlike the often used sensor raw data, our framework provides access to joint orientations which is more human-interpretable and better serves medical diagnosis.
\item We showcase that approximate {\itshape segment-to-segment orientations} are sufficiently viable to characterise sleep postures, without the need for exhaustive sensor-to-segment calibration procedures that hinder the deployment of wearable sensors in clinical or home settings.
\item We propose the use of three-dimensional (3D) {\itshape computer graphics} software to accelerate development and tune algorithms by performing an {\itshape in silico} sleep experiment before validating the methodology on human participants. Previous works often rely solely on real data, which may be hard to collect during the developmental phase.
\item We propose a novel {\itshape one-shot learning} scheme to accelerate learning of arbitrary human sleep postures with augmented observations. This eliminates the need for longitudinal data collection and labelling, which often hinder the use of wearables.
\item We built quadruple non-invasive wearable sensor modules using low-cost {\itshape off-the-shelf} components. Each module comprises two {\itshape inertial measurement units} (IMUs) to offer dual-segment tracking across the distal joint of each extremity limb.
\item We propose a metric-based approach, coupled with data visualisation, to extract quantitative and qualitative insights on posture data trends, data augmentation, and the sleep posture classification problem as a whole.
\end{itemize}


The structure of the rest of this paper is as follows. \cref{sec:relatedwork} discusses the literature relevant to the problem of human posture analysis using wearable sensors, with particular emphasis on the knowledge gap and clinical needs. In \cref{sec:methods}, we explain our methodology with reference to the proposed framework depicted in \cref{fig:systemfigure}. \cref{sec:experiments} presents the experimental design and setup, together with a description of the framework implementation. \cref{sec:resultsdiscussion} presents the evaluation results obtained and discusses the main findings. In \cref{sec:conclusions}, the paper highlights are summarised along with suggestions for future research directions. 

%
%

\section{Related Work}  \label{sec:relatedwork}
In the clinical landscape, {\itshape polysomnography} (PSG) is regarded as the gold standard clinical diagnostic tool for diagnosing sleep disorders. It involves the simultaneous recording of several parameters to evaluate two major aspects: sleep staging and physiology \cite{Vaughn2008}. Sleep stages are essential as they allow for the recovery and development of the body and the brain. Sleep staging is typically evaluated based on the brain neural activity from {\itshape electroencephalogram} signals, besides {\itshape electrooculogram} and {\itshape electromyography} which help too, particularly with the {\itshape rapid eye movement} stage \cite{Markun2020}.  The sleep physiology is necessary to assess the respiratory health, blood circulation and other functions like that of the renal and endocrine systems \cite{Colten2006}. Hence, PSG is a key tool for the diagnosis of various cardiovascular, neurologic and neuromuscular conditions, and in the evaluation of other sleep disorders such as insomnia and abnormal body movements.

Nevertheless, PSG is faced with sensing, interpretation and diagnosis challenges. To begin with, it requires patient overnight hospitalisation with $20+$ on‑body intrusive sensors which adversely affect the patient's sleep quality. As far as human posture is concerned in this work, this aspect remains premature and partially exploited in PSG. Although some PSG implementations evaluate the positional component of sleep-disordered breathing, they often come with a basic thoracic sensor that is limited to recognising only few body postures \cite{Tiotiu2011,Verbraecken2013}. An enhanced postural analysis tool would allow clinicians to have a more holistic understanding of how sleep is linked to other conditions such as musculoskeletal morbidities.


Clinicians often welcome the uprising of wearable sleep trackers \cite{Markun2020}. However, these trackers remain in the early validation phase, as it is unclear how this additional data can provide more information other than the general wellness and sleep/wake detection \cite{Razjouyan2017}. For wearable trackers, each manufacturer integrates their own proprietary algorithm and there is no widely accepted standard, unlike the case of PSG whose standard was set by the {\itshape American Academy of Sleep Medicine}.

From a motion capture perspective, the human motion analysis literature is branched into movement {\itshape quantification} and {\itshape classification} \cite{Lopez-Nava2016}. Most of the available literature belongs to the former category, and is concerned with the estimation of position and/or orientation of one or more body segments during various human activities, which are useful in sport science and film making. In contrast,  classification targets high-level interpretations or labelling of the underlying human motion/posture. Within the field of sleep posture classification, only the works on IMU-based wearables of a geo-inertial sensing modality are reviewed here as they are the most relevant for the work presented in this paper.

The available literature can be grouped according to the number of postures considered. The majority of literature consider only four standard sleep postures: {\itshape supine}, {\itshape prone}, right and left {\itshape lateral} positions. Interestingly, a single sensor attached to the chest and feeding data to a {\itshape Linear Discriminant Analysis} (LDA) classifier was sufficient to classify four sleep postures with an accuracy of $99\%$ \cite{Zhang2015}. Another study \cite{Sun2017} evaluated four classifier architectures: {\itshape Na\"ive Bayes}, {\itshape Bayesian Network}, {\itshape Decision Tree} (DT) and {\itshape Random Forest} on their performance in recognising the four postures based on statistical features extracted from an accelerometer embedded in a smartwatch. Their accuracies were found to vary between $60.3\%$ and $91.8\%$, with Random Forest being the best performer. In \cite{Eyobu2018a}, spectral features extracted from a sole upper arm sensor on a frame-by-frame basis were used to train a {\itshape Long Short-Term Memory} (LSTM) network, achieving $99\%$ in a four-posture classification problem. A recent study \cite{Alinia2020} investigated two aspects of a single-sensor sleep posture classification: (1) optimal body locations for sensor placement, and (2) the evaluation of {\itshape feature-based pattern recognition} against {\itshape deep learning} models. Given a quad-posture dataset, the comparative analysis identified the chest and either thighs as optimal body locations, and revealed comparable performance between handcrafted feature-based classifier and deep learning models. A different approach to sleep quad-posture classification was proposed in \cite{Jeon2019}, where a probabilistic state transition from one posture to another is conditioned on the inertial profile of the pose change motion. The authors defined the transitioning motion profile through the extraction of 66 different features in time and frequency domains from raw data channels sourced from triple sensors attached to the chest and wrists.

Fewer works included more sleep postures in their case studies. For a care home application \cite{BernalMonroy2020}, three classifier models were evaluated in a six-posture classification problem: (1) {\itshape k-nearest neighbours} ($k$-NN), (2) {\itshape Decision Tree} (DT), and (3) {\itshape Support Vector Machines} (SVM). Using three sensors embedded into garments (socks and T-shirt), this work adopted a pure pattern recognition approach where the authors preprocessed and extracted features from the sensory timeseries for classifier training. These pose classifications were then fed into a knowledge-based fuzzy model to automatically determine the priority level of postural changes for the prevention of pressure ulcers. The SVM was identified as the best performing classifier during pilot experiments with an accuracy of $99\%$.

A clinical study investigated the recognition of eight sleep postures using three wearable sensors placed on the forearms and chest \cite{Kwasnicki2018}. The eight postures represent minor variations of the four standard sleep postures. Using statistical features manually extracted from raw sensory data, the average four-posture classification accuracy was $99.5\%$. Notably, this figure dropped to $92.5\%$ when considering the eight minor posture variations, with the worst model accuracy hitting as low as $84.3\%$. The authors also identified battery life and large sensor size as two limitations of wearable-based sleep trackers, and provided recommendations on sensor design, packaging and data capture/transmission optimisation.

With three sensors attached to the chest and ankles, a case study explored the feasibility of classifying six to eight minor variations of the four standard sleep postures \cite{Fallmann2017}. Under different test settings, the {\itshape generalised matrix learning vector quantisation} (GMLVQ) technique was found to perform variably on 7-hour individual participant data from $58.4\%$ up to $99.8\%$. Multi-subject models were examined too and achieved a mean accuracy between $78\%$ and $83.6\%$.

More recently, we reported the classification of twelve simulated benchmark sleep postures using sparse postural cues from the four extremity limbs \cite{Elnaggar2020}. The posture dataset encloses different limb configurations common in sleep, making it more qualified for clinical use. The proposed data augmentation technique allowed for synthetically generating more postural samples and was proven to enhance the overall posture classification performance. Given a scarce dataset, the reported average classification accuracy was as high as $100\%$ using an SVM-based classifier. To emulate sensing artefacts commonly encountered in off-the-shelf sensors, mild to extreme levels of noise-based jamming were added to the testing postural samples, with the classifier showing high robustness (above $77\%$).

Some case studies investigated additional aspects of sleep as well. In \cite{Chang2018}, a smartwatch embedded with an accelerometer, microphone and illumination sensor was used to capture sleep information on the body posture, hand position and acoustic events (e.g. snores and coughs). Based on the tilt of the hand, the authors employed a 1-NN classifier to recognise the four standard body sleep postures where the similarity criterion is based on direct Euclidean distance measurement. With a 6-hour data recording per participant, the system achieved over $90\%$ accuracy in the quad-posture classification task.


Though some works do not emerge from the domain of sleep tracking, they remain relevant to intelligent wearable sensing and the analysis of human posture and movement. A smart jumpsuit with four inertial sensors on the upper arms and thighs was used for the early detection of neurodevelopmental disorders among infants through the analysis of their body postures and movements \cite{Airaksinen2020}. The authors investigated: (1) feature-based ML, and (2) end-to-end deep learning, both of which performed comparably around $95\%$. It was also shown that the quadruple sensor configuration improved the system's classification accuracy by up to $24\%$ compared to partial sensor deployment.


Another study employed a dense sensor network composed of 31 wearable sensors to classify 22 (non-sleep) body postures common in human daily activities \cite{Ohashi2018}. Despite the large number of postures considered and the high throughput of sensor data, a 1-NN classifier attained an average classification accuracy of $81\%$ using simple weighted posture attributes.


The framework proposed in this paper sits at a sweet spot between the quantification and classification branches of human motion analysis. Instead of operating on raw sensor signals, we map the sleep posture labels to the kinematic orientation space of the body's extremity limbs. Specifically, the extremity segment-to-segment relative orientations (joint angles of wrists and ankles) are regarded as primal indicators of body posture. This translates to a more explainable posture recognition algorithm and equips clinicians with better qualified diagnostic tools. With twelve sleep postures, our work goes beyond the four standard poses commonly considered in the literature. Clinically speaking, our work advances the sleep posture sensing capability which has been a main shortcoming of today's PSG systems. According to the literature, it is evident that different classification models perform comparably; from the na\"ive $k$-NN classifiers to deep learning models. Such remarkable a conclusion shall draw more attention to the data collection and treatment stages. Therefore, we leverage on a noise injection based data augmentation technique to: (1) mitigate the effect of biases present in our postures dataset, and (2) accomplish performance similar to the state-of-the-art models at a fraction of the training data. We also leverage on additional performance interpretation techniques to showcase the added value brought by data augmentation to the one-shot learning problem, while lending explainability to the reasoning behind the model.

\section{Methods}  \label{sec:methods}
This section describes the methods adopted in each stage of the proposed framework sketched in \cref{fig:systemfigure}. An overview of the framework is presented in \cref{sec:framework}. The acquisition of postural cues defining the sleep posture had been first formulated virtually through {\itshape in silico} simulations as explained in \cref{sec:virtualsleep,sec:posecharacterisation} and then similarly performed using real world data collected by the wearable sensors described in \cref{sec:sensors,sec:sensorfusion}.
The proposed postural data augmentation is described in \cref{sec:dataaugmentation}. Lastly, the model behind the posture classification is outlined in \cref{sec:postureclassification}.

\begin{figure*}[t]
    \centering
    \begin{subfigure}[b]{0.7\textwidth}
    	\centering
    	\includegraphics[width=\linewidth]{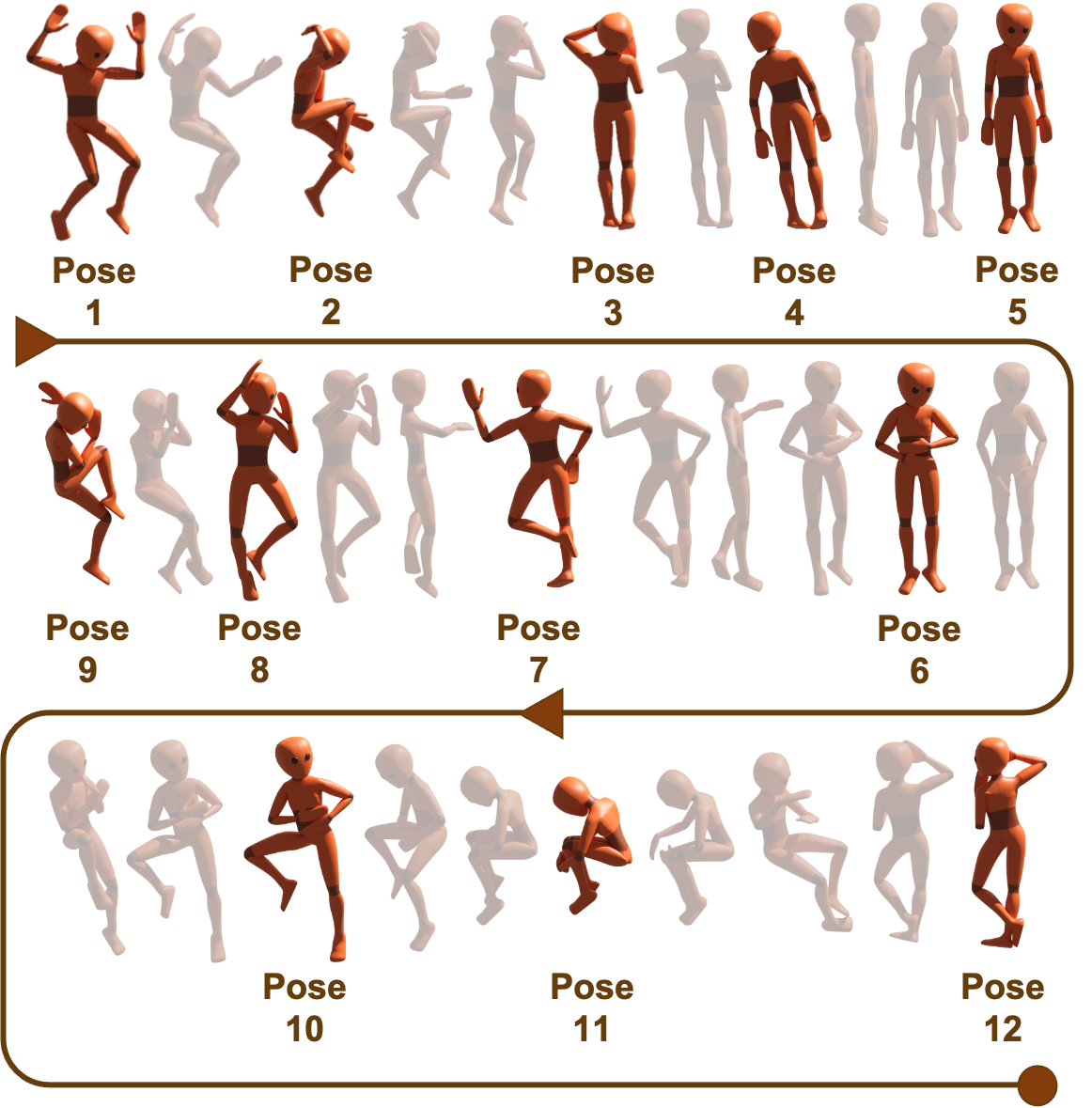}
    	\caption[Optional Caption]{}
    	\label{fig:insilicoexpa}
    \end{subfigure}
    \begin{subfigure}[b]{0.17\textwidth}
    	\centering
    	\includegraphics[width=\linewidth]{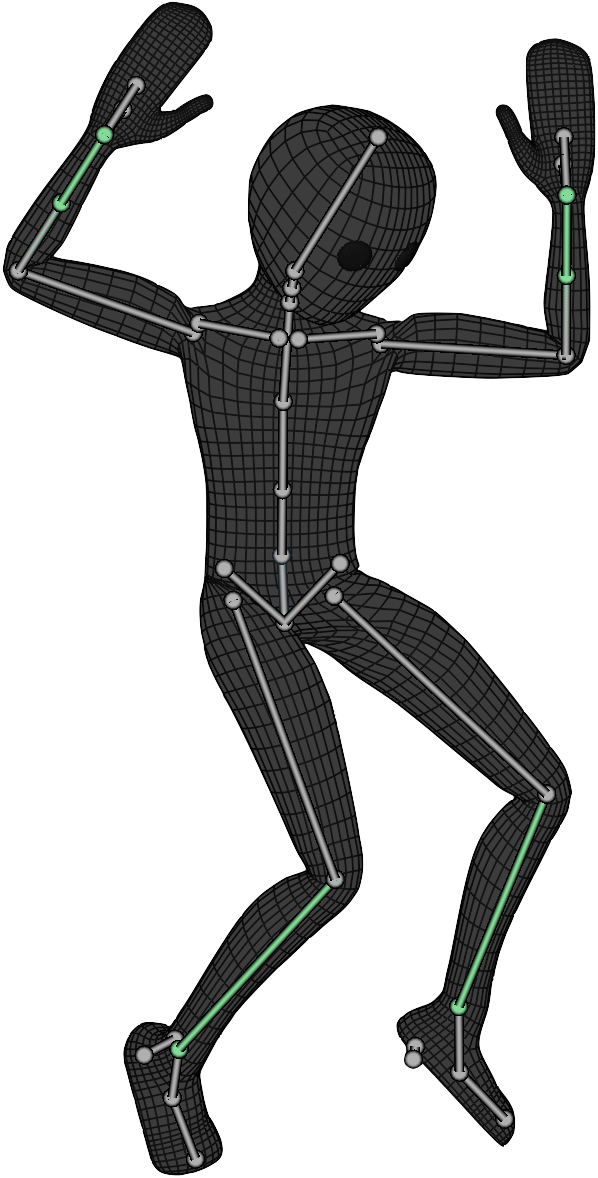}
    	\caption[Optional Caption]{}
    	\label{fig:insilicoexpb}
    \end{subfigure}
    
    \caption{In silico sleep simulation: {\bfseries (a)} motion sequence illustrating the twelve sleep postures virtually replicated in Blender$^\copyright$ , and {\bfseries (b)} anthropomorphic rig used for animating the 3D character model.}
	\label{fig:insilicoexp}
\end{figure*}

\subsection{Posture Learning Framework}  \label{sec:framework}
The proposed human posture learning framework is designed to serve as a plug-and-play system to recognise any arbitrary body posture given a single training shot for that posture.
The system comprises four wearable sensor modules and a server for sensor data acquisition, storage and analysis. In real life the system would be used as follows. Following an instruction manual or video, a subject will attach the wearable sensor modules to their wrists and ankles before sleep, then replicate a defined set of sleep postures in bed and a snapshot of sensor data is recorded at each posture. All transmitted sensor data are preprocessed to extract segment-to-segment orientations (joint angles) to be used as postural cues defining each posture (see \cref{sec:posecharacterisation}).
To avoid the need for longitudinal data collection and labelling, the single shots of preprocessed data are subsequently augmented with many more modified copies (i.e. synthetic data samples) which accelerate effective modelling of each posture. This new augmented posture dataset is sufficiently diversified, and therefore suitable for training a multi-class classifier for this particular sleep session. The patient would then sleep while the sensor data continue to be streamed to the server for sleep posture analysis. From the sequence and duration of sleep postures overnight, a clinician may be able to extract useful clinical insights. The fact that the wearable sensor modules are not taken off between the collection of the training data and real sleep data (testing data) means that sensor-to-segment misalignment is fixed throughout the recording session, thus no calibration is required.

Unlike the vast majority of literature that considers only four standard sleep postures, we showcase the scalability of the framework with twelve wide-ranging postures common in sleep. The framework has been validated in two experimental pipelines, {\itshape in silico} sleep simulation and human participant study as depicted in \cref{fig:systemfigure}.

The mechanics of the proposed framework is best explained by analogy with the flow of information in a standard pattern recognition system: {\itshape data collection}, {\itshape preprocessing}, {\itshape classifier training} and {\itshape testing}. The data collection stage involves a local server acquiring body segment orientations either from an exported sleep simulation file (see \cref{sec:virtualsleep}) or from IMU data transmitted by the wearable sensor modules (see \cref{sec:sensorfusion}).

The data preprocessing stage starts with the {\itshape body pose characterisation} step (described in \cref{sec:posecharacterisation}) to extract segment-to-segment orientations from each extremity limb to monitor joints perceived relevant from a clinical perspective. These four relative orientations serve as a simplified and human-interpretable representation of the overall body posture. The segment-to-segment orientation data are then augmented as described in  \cref{sec:dataaugmentation} to accelerate sleep posture modelling.

The use of the data augmentation step for the {\itshape in-silico} sleep simulation slightly differs from that of the {\itshape in-vivo} case. The sleep simulation provides only one observation for each posture; therefore, data augmentation is used twice to generate training and testing posture datasets, respectively, to validate the framework virtually. In contrast, the {\itshape in-vivo} session provides recordings of body postures. Therefore, data augmentation is only used to diversify the training dataset of postures, whereas the real test-labelled recordings are readily available for the purpose of framework validation in real-world. Since real timeseries are used for testing, the relative orientation data channels from the quadruple wearable sensors were synchronised.

The multi-class classification model is comprised of an ensemble of SVM binary classifiers. The classifier training and testing procedures are the same for both {\itshape in-silico} and {\itshape in-vivo} pipelines. Using the augmented posture dataset for training, the classifier model is trained to recognise the underlying sleep posture. Then, the test-labelled posture dataset is used to evaluate the performance of the pre-trained model against groundtruth labels.

\subsection{In Silico Sleep Simulation}  \label{sec:virtualsleep}
The virtual sleep simulation is built around  Blender$^\copyright$ (The Blender Foundation, Amsterdam, NL), an open-source computer graphics software for 3D modelling, animation and video rendering. A 3D human-like character model\footnote{https://cloud.blender.org/training/animation-fundamentals/5d69ab4dea6789db11ee65d1/} is virtually animated to replicate the twelve sleep postures shown in \cref{fig:insilicoexpa}. Each body posture was captured in a keyframe and transitions between keyframes were interpolated to create a motion sequence simulating sleep. We followed the standard pipeline used by digital artists for character animation. An {\itshape anthropomorphic rig} (acting as a skeleton, see \cref{fig:insilicoexpb}) is carefully aligned and bond to the character model to allow for full-body animation by posing the rig alone.

Structurally, the rig segments have root-parent-child relationships. The root is a fully unconstrained segment with six {\itshape Degrees of Freedom} (DOF) representing the translation and rotation of the rig as a whole. The root segment sits at the top of the rig's hierarchy and was chosen to be the lower spine segment. Branched off from the root segment are the kinematic chains forming the remainder of the rig (e.g. lower limbs, upper back segments, etc.). The 26 segments forming these kinematic chains follow a parent-child transformation nature in the sense that rotation or translation of a parent segment,  affects the pose of all its subsequent child segments, but not vice versa.

The complete definition of the body posture, $\bm{B}$, is defined by two components: (1) the combined position and orientation of the root segment $\bm{\rho} \in \mathbb{R}^6$, and (2) the rotations vector $\bm{\alpha} \in \mathbb{R}^n$ of the remaining 26 rig segments. The $\bm{\alpha}$ vector contains the angular displacements about $n$ active rotational axes of all body segments, depending on the joint definitions ({\itshape ball}, {\itshape saddle}, or {\itshape hinge} joints). To allow for body posture tracking, Blender$^\copyright$ automatically assigns right-handed 3D coordinate systems $\{B_i \ | \ i \in \mathbb{Z} : 1 \leq i \leq D\}$ to all $D$ segments anchored at their respective parent joints' active centres of rotation. Given an arbitrary $j^{\text{th}}$ segment, its pose can be referred to a reference coordinate system, $R$, using the {\itshape shape forward kinematics map}
\begin{align}
    \prescript{R}{}{\bm T}_{j} &= {\bm T}_{1}\ {\bm T}_{2}\ ...\ {\bm T}_{j}\prescript{R}{}{\bm T}_{j}(0) \nonumber\\
    &= \left(\prod_{i=1}^{j}\ {\bm T}_{i}\right)\prescript{R}{}{\bm T}_{j}(0)
    \label{eq:1}
\end{align}
where ${\bm T}_{i} \in \mathbb{R}^{4\times4}$ denotes the homogeneous transformation matrix describing the rotations and translations of $B_i$, and $\prescript{R}{}{\bm T}_{j}(0)$ represents the initial postural offset between coordinate systems $j$ and $R$ after the rig binding process is completed.
This map allows for the calculation of the net transformation of a child segment by combining all hierarchical transformations from parent segments. In Blender$^\copyright$, $R$ is often the global coordinate system of the 3D viewport, and thus the offset term for each segment is known {\itshape a priori}.

The terms ${\bm T}_{i}$ can be exported with the animation rendered video as a {\itshape BioVision Hierarchy} (BVH) file containing these hierarchical transformations, quantified by the angular and translational displacements of each segment at each frame, and the fixed segment-to-segment positional offsets.

\subsection{Characterisation of Body Posture}  \label{sec:posecharacterisation}
In principle, a posture is defined by the complete set of joint angles of all body segments, which is an unrealistic measurement and computing challenge for wearable sensing. Therefore, what we refer to as ``{\itshape body pose characterisation}'' is the selection of relatively few joint angles that are practically measurable and, at the same time, allow sleeping postures to be classified. The definition of the sleep posture is important since the outcome of this step will have a strong impact on the selection of effective techniques for collecting and analysing data.

The challenge with the body posture is that its study follows a dual nature of {\itshape parametric} and {\itshape subjective} aspects. It is parametric in the sense that measurements of some modality are required for algorithms to use in decision making. This is considered the mainstream direction of the available literature as covered in \cref{sec:relatedwork}. Subjectivity is more related to the human perception of the sleep posture, which varies from a person to another. For example, in \cite{Airaksinen2020}, multiple human annotators were found to disagree in labelling postures captured in video. Having said that, subjectivity is not completely disjointed from measurements; humans need high-level information in some form (e.g. images, or numbers), but not raw sensor readings. The subjective element in posture classification is useful since posture measurement variability within some constraint should be permissible. In this paper, we exploit the parametric-subjective nature for posture characterisation (herein this section), and augmentation (see \cref{sec:dataaugmentation}).

\begin{figure}[b]
    \centering
    \includegraphics[width=\columnwidth]{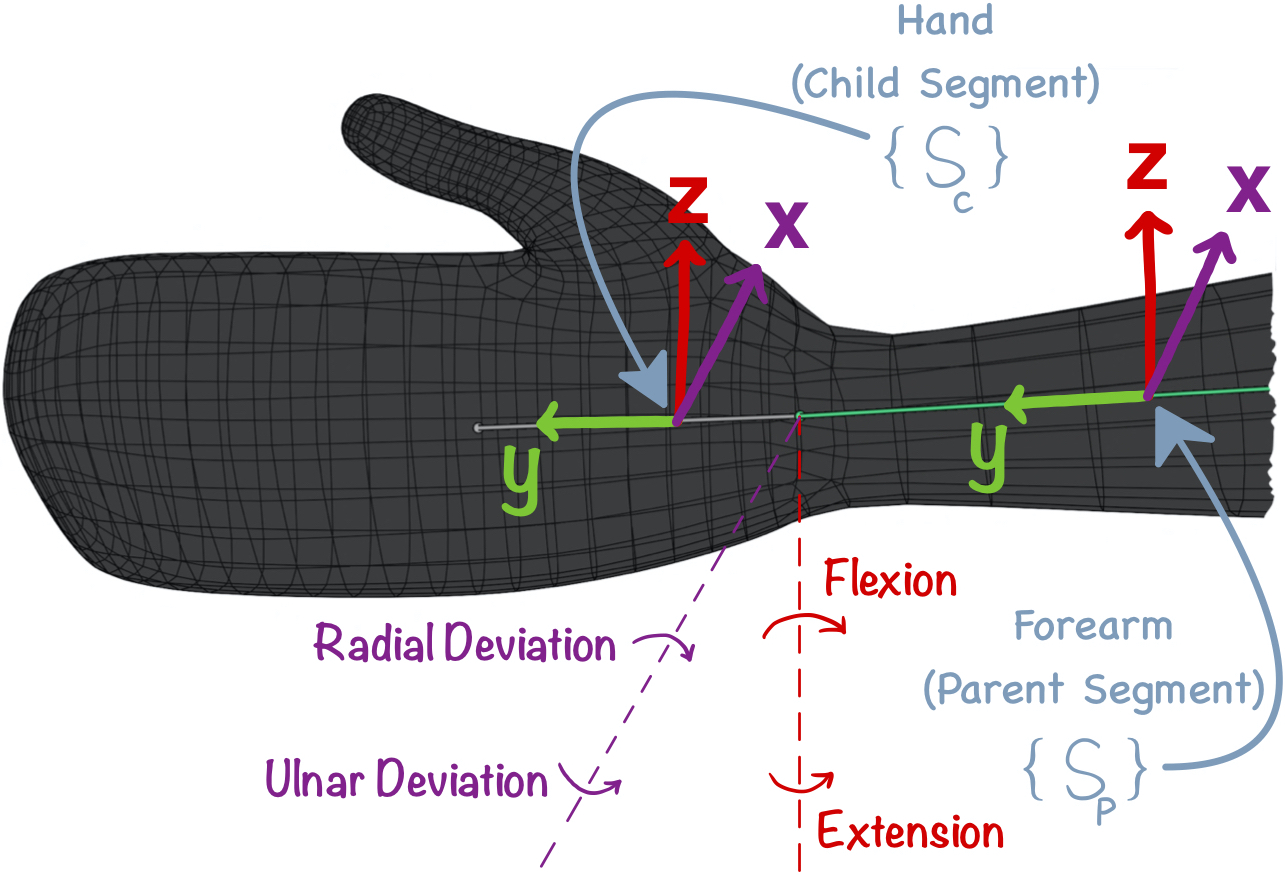}
    \caption{The kinematic definition of the wrist joint.}
    \label{fig:annotatedwrist}
\end{figure}


To reach a good compromise between the numerical and perceptual reasoning behind postural analysis, we propose the use of segment-to-segment relative orientations at the four extremity limbs (wrists and ankles) as primal indicators of the sleep posture. This provides clinicians with an advantageous access to human-interpretable and simplified posture definition alongside the output pose labels. The choice of ankle and wrist joints stems from their strong connection with various sleep-related pathologies, such as {\itshape ankle osteoarthritis} \cite{Mobasheri2016} and {\itshape carpal tunnel syndrome} \cite{McCabe2007}.
Such intuitive postural information is envisaged to bring clinicians more comprehensibility of the posture classification algorithm, making it a better fit as a future medical diagnostic tool.

Segment-to-segment relative orientations represent the rotational component of the local joint transformation linking a child segment to its parent. For illustration, \cref{fig:annotatedwrist} shows the right wrist joint and the coordinate systems $S_p$ and $S_c$ of, respectively, the forearm (parent segment) and the hand (child segment). The wrist is a {\itshape condyloid synovial} (or saddle) joint allowing only two motions: {\itshape flexion/extension} and {\itshape ulnar/radial deviation}. Based on this definition, the hand-to-forearm rotation matrix, $\prescript{S_p}{}{\bm R}_{S_c}$, can be formulated as
\begin{equation}
    \prescript{S_p}{}{\bm R}\prescript{}{S_c} = {\bm R}\prescript{}{z}\ {\bm R}\prescript{}{y}\ {\bm R}\prescript{}{x}\
    \label{eq:2}
\end{equation}
where, in the case of the wrist joint, ${\bm R}\prescript{}{z}\ $and ${\bm R}\prescript{}{x}\ $represent the rotations of the flexion/extension and ulnar/radial deviation respectively. The wrist pronation/supination originates from the elbow joint, hence ${\bm R}\prescript{}{y}\ $is ideally an identity matrix.

For the {\itshape in silico} sleep simulation, segment-to-segment orientations can be derived from the skeleton hierarchical transformations provided in the BVH file exported from Blender$^\copyright$. Indeed, using \cref{eq:1}, the relative transformation between the parent and child segments can be obtained as

\begin{equation}
    \prescript{S_p}{}{\bm T}_{S_c} = \left(\prescript{R}{}{\bm T}_{S_p}\right)^{T} \prescript{R}{}{\bm T}_{S_c}
    \label{eq:3}
\end{equation}

The rotational component of the local transformation across the extremity limb distal joint can then be extracted as

\begin{equation}
    \prescript{S_p}{}{\bm R}\prescript{}{S_c} =\ {\bm Q}\ \prescript{S_p}{}{\bm T}_{S_c}\ {\bm Q}^T
    \label{eq:4}
\end{equation}
where
\begin{equation*}
    \bm{Q} = 
    \begin{bmatrix}
        \bm{I}_{3\times3} & \bm{0}_{3\times1}
    \end{bmatrix}
    \label{eq:4_1}
\end{equation*}

Similar kinematic definitions are made for the lower extremity limbs. In this case, the local transformation of the ankle joint can be monitored by tracking both the shin and foot segments, with the allowable ankle motions being the {\itshape inversion/eversion} and {\itshape plantar/dorsi-flexion}.

The BVH articulated body representation is often expressed in Euler angle-based rotation matrices as defined in \cref{eq:2}. However, for the purpose of this study  $\prescript{S_p}{}{\bm R}\prescript{}{S_c}\ $ was converted to its equivalent quaternion form $\prescript{_{S_p}}{}{\bm q}_{_{S_c}}$ to obtain a more concise and numerically stable representation. Thus, the pose characterisation vector ${\bm \rchi}^v$ for the virtual character model is defined as
\begin{equation}
	{\bm \rchi}^v =
	\begin{bmatrix}
        \prescript{_{S_p}}{}{\bm q}_{_{S_c}}^{\mathbfcal{J}(1)} & \prescript{_{S_p}}{}{\bm q}_{_{S_c}}^{\mathbfcal{J}(2)} & \prescript{_{S_p}}{}{\bm q}_{_{S_c}}^{\mathbfcal{J}(3)} & \prescript{_{S_p}}{}{\bm q}_{_{S_c}}^{\mathbfcal{J}(4)}
    \end{bmatrix}
	\label{eq:5}
\end{equation}
where $\mathbfcal{J} = \{\text{\itshape right wrist}, \text{\itshape left wrist},\text{\itshape right ankle}, \text{\itshape left ankle}\}$ denotes the set of four distal joints of the four extremity limbs.


It is worth noting that the pose characterisation framework proposed in this paper does not utilise any calibration poses and exploits segment-to-segment orientations, making the approach more meaningful clinically. This goes beyond the approach previously presented by the authors in \cite{Elnaggar2020} which required a reference calibration T-pose, and offered tracking of the child segment alone.

\begin{figure*}[t]
    \centering
    \begin{subfigure}[b]{0.2\textwidth}
    	\centering
    	\includegraphics[width=\linewidth]{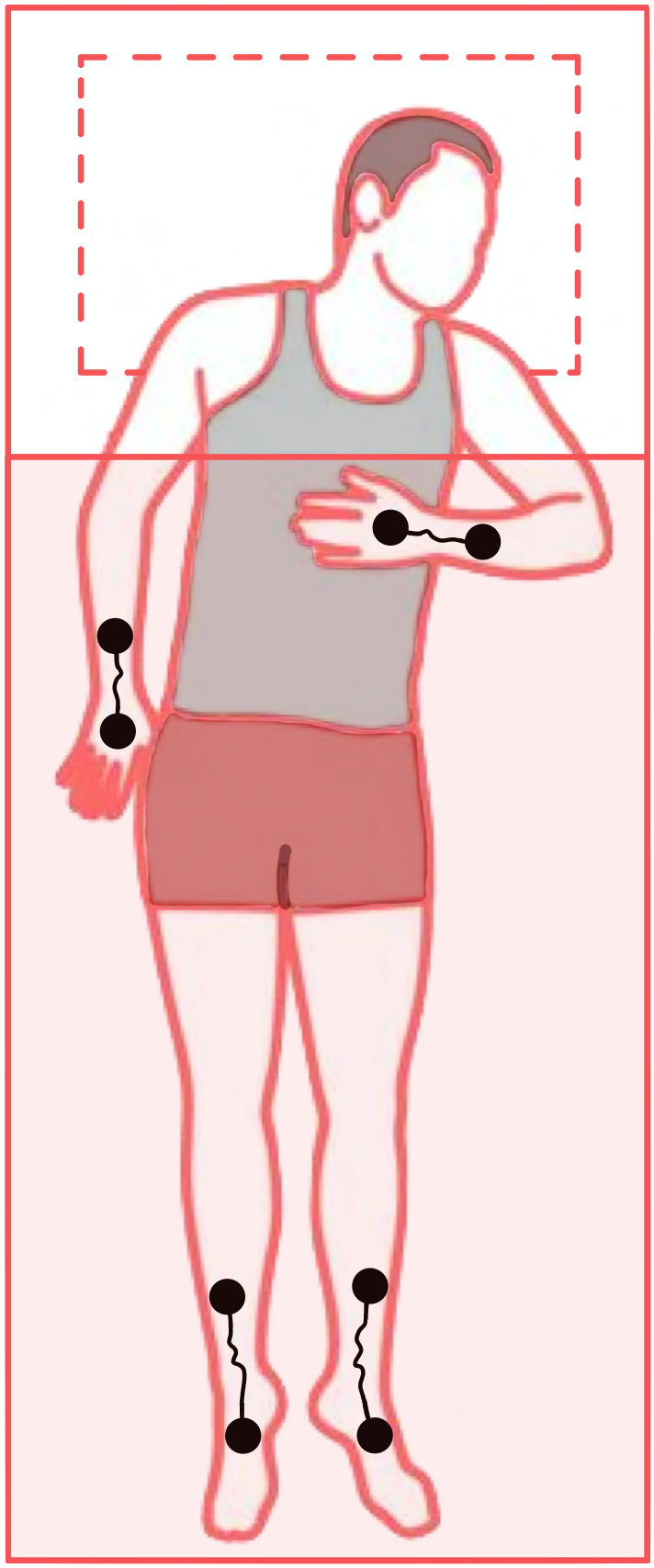}
    	\caption[Optional Caption]{}
    	\label{fig:sensorplacement}
    \end{subfigure}
    \hspace{0.2\textwidth}
    \begin{subfigure}[b]{0.36\textwidth}
    	\centering
    	\includegraphics[width=\linewidth]{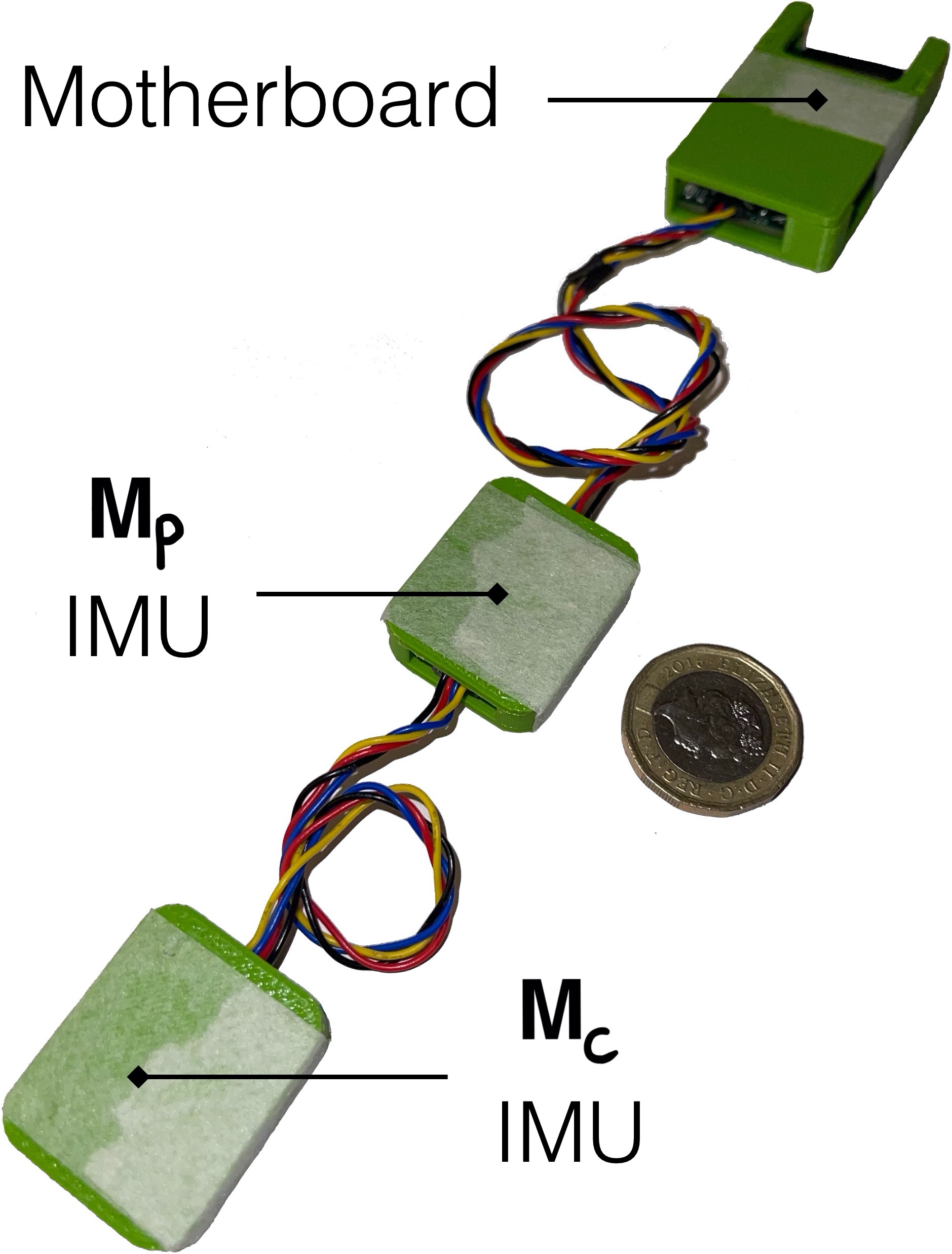}
    	\caption[Optional Caption]{}
    	\label{fig:sensormodule}
    \end{subfigure}
    
    \caption{The real-world experimental setup: {\bfseries (a)} placement of the wearable posture sensors on the four extremity limbs; each two black-filled circles represent one complete limb tracker, and {\bfseries (b)} annotated illustration of the wearable sensor module.}
	\label{fig:realexp}
\end{figure*}

\subsection{Wearable Posture Sensors}  \label{sec:sensors}
Some form of wearable technology is required to track segment-to-segment orientation of the four distal joint angles. The main requirements for such wearable technology include: (1) multi-segment orientation tracking, (2) compact size to not compromise sleep quality, and (3) low cost to make it affordable to the public health sector. Predominantly, reported works on human motion analysis involving several body segments simply employ multiple standalone wearable sensors; one for each segment. In this work, we opt to design a custom-made sensor module (shown in \cref{fig:sensormodule}) with dual-segment tracking capability, empowered by two embedded BNO055 IMU sensors from Bosch Sensortec$^\copyright$ (Bosch Sensortec GmbH, Reutlingen, DE). Both IMU sensors are managed by a single ESP32-WROOM-32D {\itshape microcontroller} from Espressif Systems$^\copyright$ (Espressif Systems Shanghai Co Ltd, Shanghai, CN) featuring Bluetooth connectivity for wireless data transmission. At about $6\ \text{cubic centimeters}$ in volume for each IMU case, the sensor module is sufficiently slim and small for wearability during sleep. Moreover, all the electronic components used in this design are commercially available, with a total low cost of approximately GBP 100. \cref{fig:sensorplacement} illustrates the on-body placement of these sensor modules such that the parent and child IMU sensors are mounted on the last two segments of each extremity limb.

\subsection{Intra- and Inter-Sensor Fusion}  \label{sec:sensorfusion}
A {\itshape sensor fusion} algorithm is needed to estimate the attitude of each IMU sensor (intra-sensor fusion), that is, a function of the body segment it is mounted on. Afterwards, a pose characterisation framework is employed in a similar way to that described in \cref{sec:posecharacterisation}. To this end, an inter-sensor fusion step is applied to fuse the two absolute IMU orientations for each wearable sensor module into one segment-to-segment orientation.

To compensate for the drift inherent to the IMU heading estimates, readings from the magnetometer, embedded in the IMU, was exploited to provide a stable estimate of the orientation. Herein, the {\itshape Madgwick filter} \cite{Madgwick2011} is employed for fusing the geo-inertial measurements from the IMU sensor, thanks to its orientation tracking robustness and successful deployment in human motion analysis research \cite{Xiao2018}. Furthermore, the optimisation procedure of the filter is of low computational cost and takes place in the quaternion space, allowing for online and singularity-free attitude estimation. As formulated in \cref{eq:6}, the filter first carries out a {\itshape vector observation} step which involves iteratively searching for an optimal orientation estimate, $\prescript{_M}{}{\hat{\bm q}}_{_E},$ defined from the IMU frame $M$ to the Earth frame $E$. The validity criterion for the orientation estimate depends on how well it aligns a sensor-measured field vector $\prescript{_M}{}{\bm s} =
	    \begin{bmatrix}
			0\quad s_x\quad s_y\quad s_z
		\end{bmatrix}$
with some Earth-referenced geophysical quantity $\prescript{_E}{}{\bm r} =
	    \begin{bmatrix}
			0\quad r_x\quad r_y\quad r_z
		\end{bmatrix}$.
\begin{equation}
	\prescript{_M}{}{\bm q}_{_E} =
		\min_{ \prescript{_M}{}{\bm q}_{_E} \in \mathbb{R}^4} \quad
		{\bm f}\left(
		\prescript{_M}{}{\bm q}_{_E}, \prescript{_E}{}{\bm r}, \prescript{_M}{}{\bm s}
		\right)
	\label{eq:6}
\end{equation}
such that
\begin{equation*}
	{\bm f}\left(
	\prescript{_M}{}{\bm q}_{_E}, \prescript{_E}{}{\bm r}, \prescript{_M}{}{\bm s}
	\right)
	= \prescript{_M}{}{\bm q}_{_E}^{*} \otimes \prescript{_E}{}{\bm r} \otimes \prescript{_M}{}{\bm q}_{_E} - \prescript{_M}{}{\bm s}
	\label{eq:6_1}
\end{equation*}
where the operator $\otimes$ denotes quaternion multiplication.

The filter then uses the Jacobian matrix of the vector objective function to determine its gradient $\nabla {\bm f}$, which is later used to define the normalised quaternion estimation error $\prescript{_M}{}{\bm q}_{_E}^{\epsilon}$ at time index $t_k$
\begin{equation}
	\prescript{_M}{}{\bm q}_{_E}^{\epsilon}(t_k) =
	    \left.\frac{\nabla{\bm f}}{\norm{\nabla{\bm f}}}\right\vert_{t_k}
	\label{eq:7}
\end{equation}

Geophysical vector observation alone provides a sluggish orientation estimate since it is a memoryless framework and is highly susceptible to sensor noise. As shown in \cref{eq:8,eq:9}, the Madgwick filter produces a smoother orientation estimate through numerically integrating a reliable orientation rate estimate $\prescript{_M}{}{\dot{\bm q}}_{_E}$ at each descent update step. The orientation rate is the outcome of fusing $\prescript{_M}{}{\bm q}_{_E}^{\epsilon}$, weighted by a hyperparameter $\left(\beta << 1\right)$, with the rate of orientation change $\prescript{_M}{}{\dot{\bm q}}_{_E}^{\omega}$ derived from the gyroscope measurement vector $\prescript{_M}{}{\bm \omega} =
	    \begin{bmatrix}
			0\quad \omega_x\quad \omega_y\quad \omega_z
		\end{bmatrix}$.
\begin{equation}
	\prescript{_M}{}{\hat{\bm q}}_{_E}(t_k) =
	    \prescript{_M}{}{\hat{\bm q}}_{_E}(t_{k-1}) + \prescript{_M}{}{\dot{\bm q}}_{_E}(t_k) \cdot \Delta t_k
	\label{eq:8}
\end{equation}
\begin{equation}
	\prescript{_M}{}{\dot{\bm q}}_{_E}(t_k) =
	    \prescript{_M}{}{\dot{\bm q}}_{_E}^{\omega}(t_k) - \beta \prescript{_M}{}{\bm q}_{_E}^{\epsilon}(t_k)
	\label{eq:9}
\end{equation}
where
\begin{equation*}
	\prescript{_M}{}{\dot{\bm q}}_{_E}^{\omega}(t_k) =
	    \frac{1}{2} \prescript{_M}{}{\hat{\bm q}}_{_E}(t_{k-1}) \otimes \prescript{_M}{}{\bm \omega}(t_k)
	\label{eq:9_1}
\end{equation*}

\begin{figure}[!b]
    \centering
    \includegraphics[width=\columnwidth]{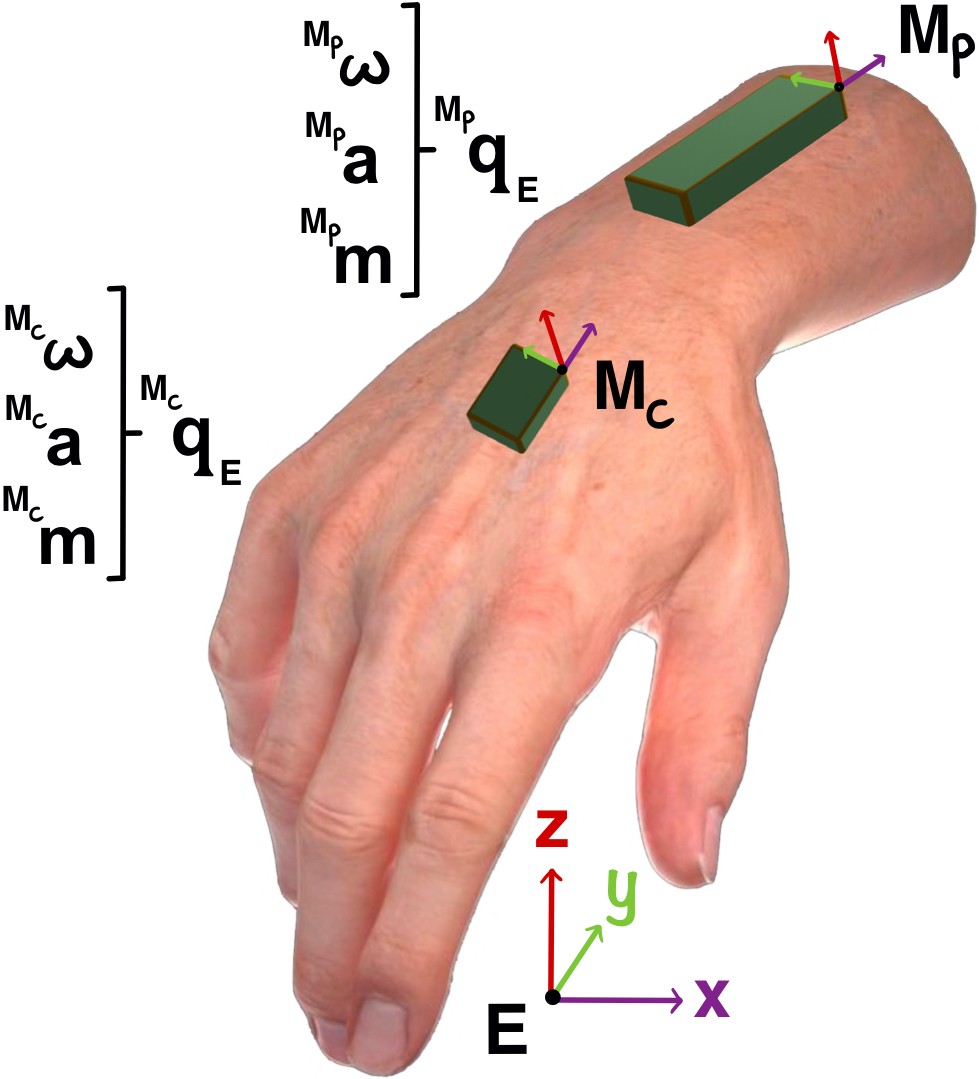}
    \caption{An illustration of wearable posture sensing for the upper limb. By applying Madgwick filtering to the angular velocity, acceleration and magnetic measurements from the IMUs, the absolute orientations $\prescript{_{M_p}}{}{\bm q}_{_E}$ and $\prescript{_{M_c}}{}{\bm q}_{_E}$ of each IMU can be estimated.}
    \label{fig:handsensors}
\end{figure}

As depicted in \cref{fig:handsensors}, the two IMUs $M_p$ and $M_c$ built in each wearable sensor modules are attached to the two most distal segments of each limb respectively. Leveraging on the sensor fusion algorithm outlined above, the absolute orientations of both IMUs are first estimated, and then fused to determine the IMU-to-IMU orientation $\prescript{_{M_p}}{}{\bm q}_{_{M_c}}$ as
\begin{equation}
	\prescript{_{M_p}}{}{\bm q}_{_{M_c}} = \prescript{_E}{}{\bm q}_{_{M_c}} \otimes \prescript{_{M_p}}{}{\bm q}_{_E} = \prescript{_{M_c}}{}{\bm q}_{_E}^* \otimes \prescript{_{M_p}}{}{\bm q}_{_E}
	\label{eq:10}
\end{equation}
This quaternion is computed for each extremity limb to approximately measure the underlying segment-to-segment orientation. Unlike works that prerequisite IMU-to-segment misalignment calibration \cite{Nazarahari2019,Zimmermann2018}, the proposed framework instead aims at fast posture classification using approximate segment orientations. Fortunately, this serves the feasibility of the proposed system since it is impractical to calibrate eight IMUs for each use.

Similar to \cref{eq:5}, the pose characterisation vector ${\bm \rchi}^w$ based on wearable sensor data is defined as
\begin{equation}
	{\bm \rchi}^w =
	\begin{bmatrix}
        \prescript{_{M_p}}{}{\bm q}_{_{M_c}}^{\mathbfcal{J}(1)} & \prescript{_{M_p}}{}{\bm q}_{_{M_c}}^{\mathbfcal{J}(2)} & \prescript{_{M_p}}{}{\bm q}_{_{M_c}}^{\mathbfcal{J}(3)} & \prescript{_{M_p}}{}{\bm q}_{_{M_c}}^{\mathbfcal{J}(4)}
    \end{bmatrix}
	\label{eq:11}
\end{equation}

\subsection{Postural Data Augmentation}  \label{sec:dataaugmentation}
This section embarks on the stage of data preprocessing where segment-to-segment orientations are augmented to create a larger dataset better suited for the ML algorithm (see \cref{sec:postureclassification}) for pose classification.
To begin with, let a generic variable ${\bm \rchi}$ be defined as either ${\bm \rchi}^v$ or ${\bm \rchi}^w$, depending on whether posture tracking is taking place {\itshape in silico} or real world. Next, we define a collective pose characterisation vector ${\bm \Psi}$ that brings together all the twelve sleep postures
\begin{equation}
	{\bm \Psi} =
	\begin{bmatrix}
        {\bm \rchi}_1 & {\bm \rchi}_2 & ... & {\bm \rchi}_{12}
    \end{bmatrix}
	\label{eq:12}
\end{equation}
where ${\bm \rchi}_j$ corresponds to the $j^{\text{th}}$ sleep posture.


Based on this definition, ${\bm \Psi}$ resembles a reference dictionary containing postural cues belonging to the sleep postures included in the presented case study. In practice, with such single-observation definitions of postures, over-fitting and poor generalisation are clearly unavoidable outcomes for any classifier regardless of its type. As covered in \cref{sec:relatedwork}, related works record extended sensor data timeseries for each posture which sometimes reach several hours or nights worth of training data. Eventually, extended data collection translates to a higher cost of manual data labelling. Moreover, each subject may have slightly different sleep postures that are of interest to clinicians; training data collection and labelling should then be repeated for each subject. This would clearly be an obstacle for clinical use of wearable-based sleep monitoring solutions.

\begin{figure*}[!t]
    \centering
    \begin{subfigure}[b]{0.49\textwidth}
    	\centering
    	\includegraphics[width=\linewidth]{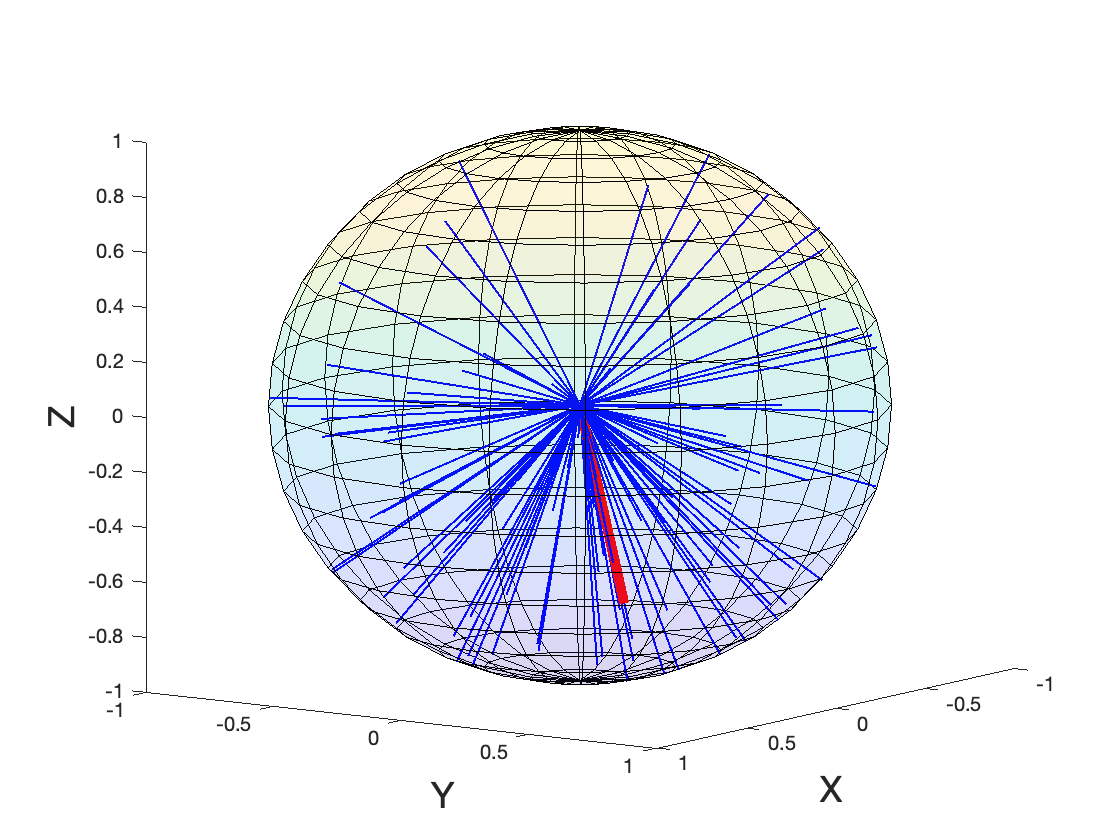}
    	\caption[Optional Caption]{}
    	\label{fig:naiveaxes}
    \end{subfigure}
    \begin{subfigure}[b]{0.49\textwidth}
    	\centering
    	\renewcommand{\thesubfigure}{c}
    	\includegraphics[width=\linewidth]{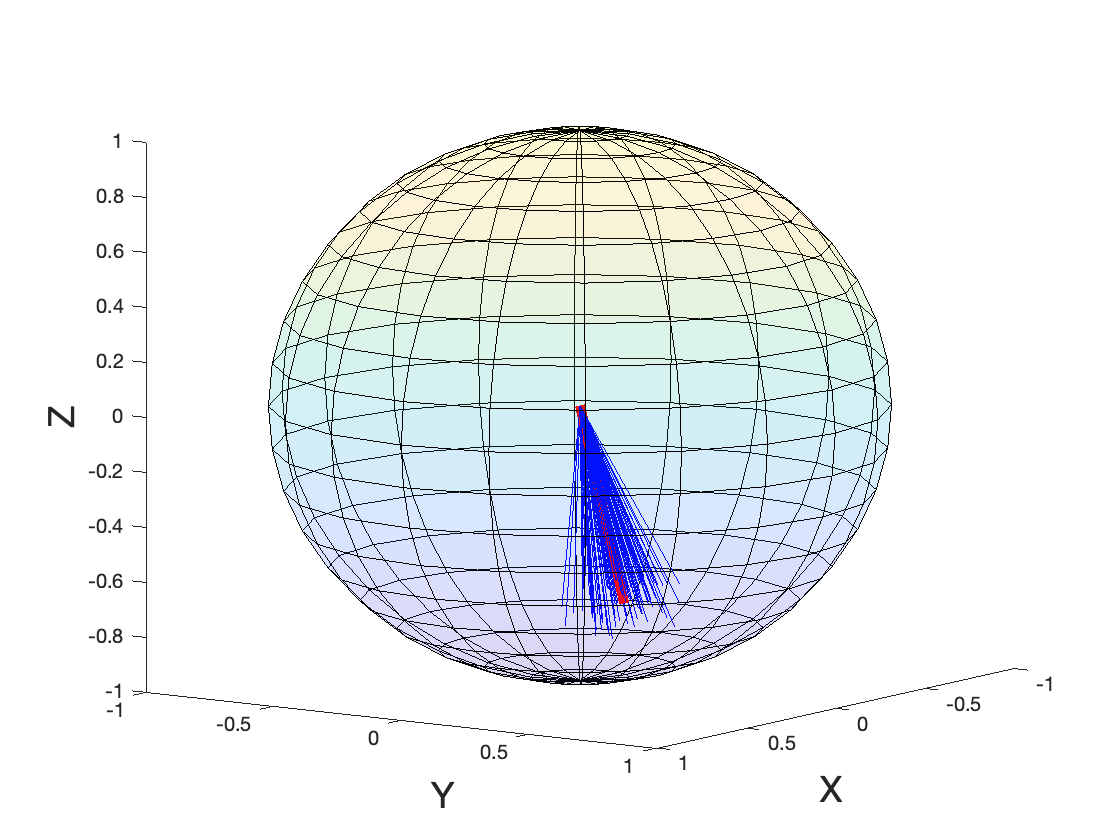}
    	\caption[Optional Caption]{}
    	\label{fig:ouraxes}
    \end{subfigure}
    \par
    \begin{subfigure}[b]{0.49\textwidth}
    	\centering
    	\renewcommand{\thesubfigure}{b}
    	\includegraphics[width=\linewidth]{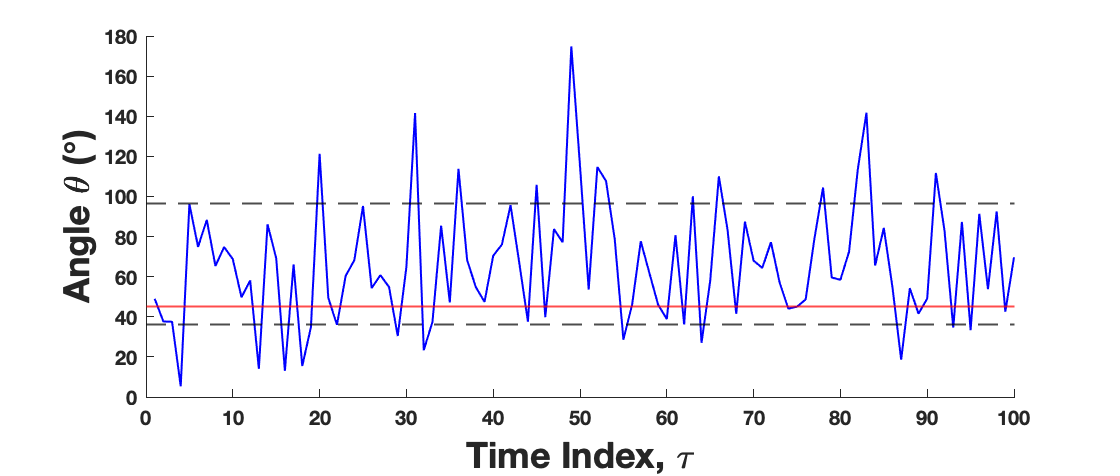}
    	\caption[Optional Caption]{}
    	\label{fig:naiveangle}
    \end{subfigure}
    \begin{subfigure}[b]{0.49\textwidth}
    	\centering
    	\includegraphics[width=\linewidth]{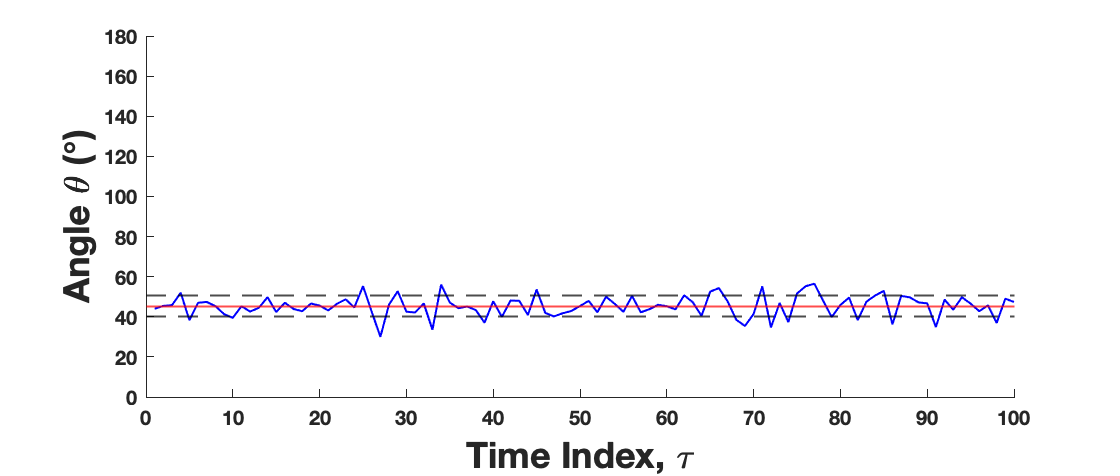}
    	\caption[Optional Caption]{}
    	\label{fig:ourangle}
    \end{subfigure}
    \caption{Postural data augmentation results ($N=100$): {\bfseries (a)} augmented axes and {\bfseries (b)} angles of rotation after injecting Gaussian noise (variance = $0.1$) to a quaternion; {\bfseries (c)} augmented axes and {\bfseries (d)} angles of rotation after injecting Gaussian noise ($\sigma_\phi=\sigma_\theta=30\degree$) to the axis-angle based orientation (proposed method). Blue-coloured data are synthetically generated, whereas the input reference axis-angle orientation is in red. The black dashed lines in {\bfseries (b)} and {\bfseries (d)} represent the sample standard deviation of the angle timeseries.}
	\label{fig:dataaugmentation}
\end{figure*}

In this work, {\itshape data augmentation} is a key preprocessing step to make a trade-off between the cost of data collection and timeseries classification. It is essentially needed in applications where only scarce \cite{Olson2018} or class-imbalanced \cite{Blagus2013} datasets are available. Another possible use of data augmentation is to obtain a more capable ML model through enhancing the quantity and quality of the training data by deliberately introducing synthetic samples. When assigned correct labels, synthetic data allows the ML model to explore regions of the input space dismissed in the real training dataset. This leads to expanding the decision boundary of the model, thus lowering the risk of over-fitting \cite{Shorten2019}.

Several families of data augmentation techniques are extensively reviewed in \cite{Wen2020,Iwana2021} which include, but not limited to, {\itshape pattern mixing}, {\itshape signal decomposition} and {\itshape generative neural networks}. However, these techniques prerequisite medium to large timeseries datasets, thus directly applying any of them to our single ``snapshots'' of postures is irrational. To address this one-shot learning problem, we propose a {\itshape noise injection} based data augmentation approach to facilitate timeseries generation having only provided a single observation of each posture. The addition of artificial noise helps in overcoming the scarcity and bias issues present in the training data, and provide a good compromise between the parametric and subjective aspects of the human pose definition. Another advantage of noise injection is that the noise generation process can be easily modelled which means that the data augmentation is both editable and invertible. Artificially noised datasets reportedly led to increased robustness to sensor noise and improved classification performance in real world applications, including construction equipment activity recognition \cite{Rashid2019a} and meteorological sensor data processing \cite{Arslan2019}.

\begin{figure}[!p]
    \centering
    \begin{tikzpicture}
        \node[anchor=south west,inner sep=0] (image) at (0,0) {\includegraphics[width=\columnwidth]{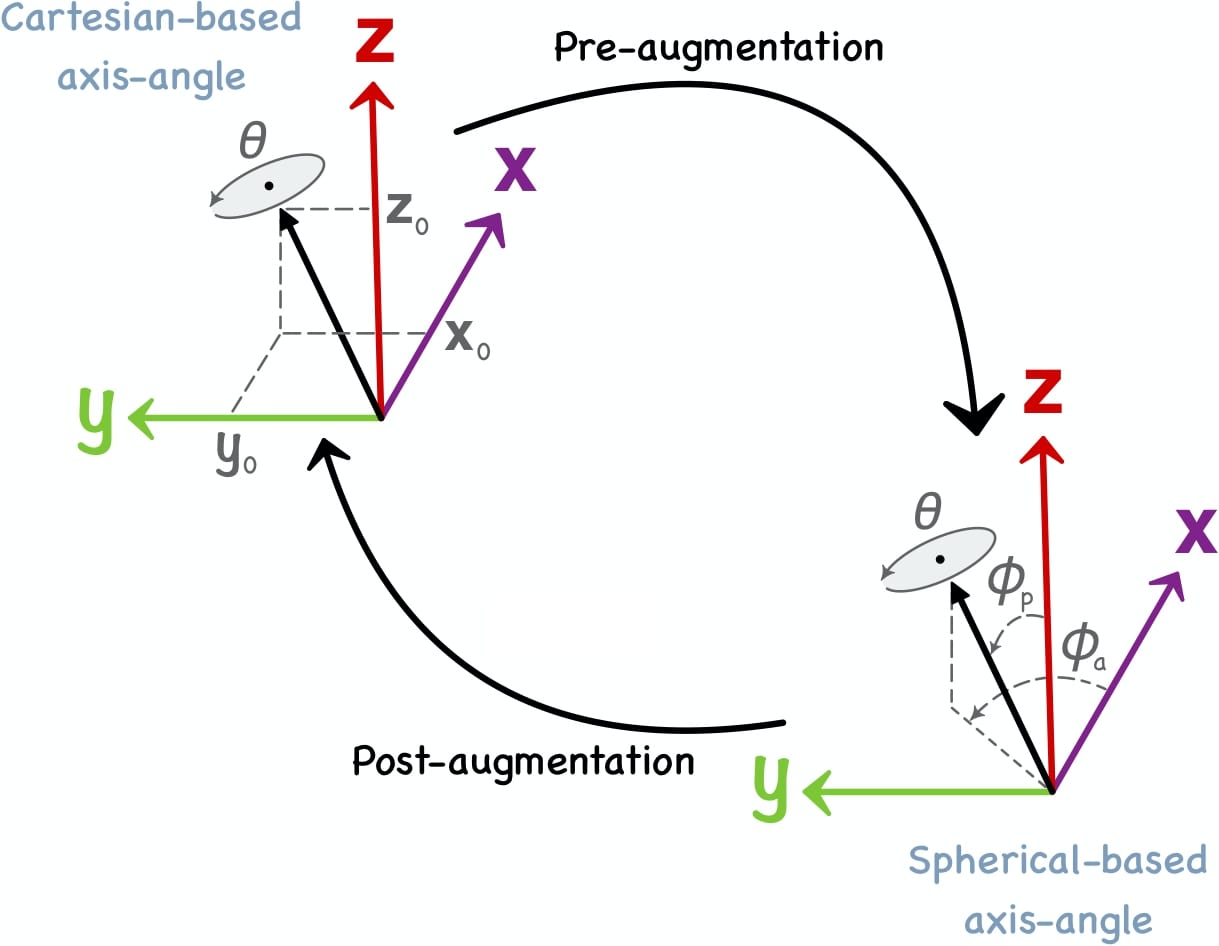}};
        \begin{scope}[x={(image.south east)},y={(image.north west)}]
            \node[anchor=center] at (0.45,0.35) {\cref{eq:13}};
        \end{scope}
    \end{tikzpicture}
    \caption{Annotated visualisation of Cartesian- and spherical-based axis-angle representations.}
    \label{fig:sphericaltrick}
\end{figure}

Nonetheless, simple addition of noise to a quaternion leads to a chaotic data augmentation process with nonsensical synthetic samples as illustrated in \cref{fig:naiveaxes,fig:naiveangle}. Therefore, to generate near-realistic postural data, the quaternion-based pose descriptor ${\bm \rchi}$ is first converted into its corresponding  {\itshape axis-angle representation} ${\bm x}$. As shown in \cref{fig:sphericaltrick}, the axis of rotation is defined in the singularity-free Cartesian space, while the augmentation step is performed in an intermediate spherical coordinate system to obtain a more homogeneous augmented dataset. In particular, ${\bm x}$ is defined as


\begin{equation}
    \begin{split}
        {\bm x} &= \begin{bmatrix}
            {\bm G}(\phi_p^{\mathbfcal{J}(1)}, \phi_a^{\mathbfcal{J}(1)}) \cdot \prescript{p}{}{\bm e}_{c}^{\mathbfcal{J}(1)} \vspace{0.1cm}\\
            {\bm G}(\phi_p^{\mathbfcal{J}(2)}, \phi_a^{\mathbfcal{J}(2)}) \cdot \prescript{p}{}{\bm e}_{c}^{\mathbfcal{J}(2)} \vspace{0.1cm}\\
            {\bm G}(\phi_p^{\mathbfcal{J}(3)}, \phi_a^{\mathbfcal{J}(3)}) \cdot \prescript{p}{}{\bm e}_{c}^{\mathbfcal{J}(3)} \vspace{0.1cm}\\
            {\bm G}(\phi_p^{\mathbfcal{J}(4)}, \phi_a^{\mathbfcal{J}(4)}) \cdot \prescript{p}{}{\bm e}_{c}^{\mathbfcal{J}(4)}
        \end{bmatrix} +
    	\begin{bmatrix}
            {\bm g}(\phi_p^{\mathbfcal{J}(1)}, \phi_a^{\mathbfcal{J}(1)}) \vspace{0.1cm}\\
            {\bm g}(\phi_p^{\mathbfcal{J}(2)}, \phi_a^{\mathbfcal{J}(2)}) \vspace{0.1cm}\\
            {\bm g}(\phi_p^{\mathbfcal{J}(3)}, \phi_a^{\mathbfcal{J}(3)}) \vspace{0.1cm}\\
            {\bm g}(\phi_p^{\mathbfcal{J}(4)}, \phi_a^{\mathbfcal{J}(4)})
        \end{bmatrix}\\
        &=
        \begin{bmatrix}
            {\bm x}^{\mathbfcal{J}(1)} & {\bm x}^{\mathbfcal{J}(2)} & {\bm x}^{\mathbfcal{J}(3)} & {\bm x}^{\mathbfcal{J}(4)}
        \end{bmatrix}^T
    \end{split}
	\label{eq:13}
\end{equation}
such that ${\bm G}(\cdot) \in \mathbb{R}^{4\times3}$ and ${\bm g}(\cdot) \in \mathbb{R}^{4\times1}$ denote parametric matrix and vector, respectively, used to transform the axis-angle representation from the spherical space to the Cartesian space, and a generic $\prescript{p}{}{\bm e}_{c}^{\mathbfcal{J}(j)}$ is defined as
\begin{equation}
	\prescript{p}{}{\bm e}_{c}^{\mathbfcal{J}(j)} = \underbrace{\phi_p^{\mathbfcal{J}(j)}\ {\bm {\hat e}}_{_1} + \phi_a^{\mathbfcal{J}(j)}\ {\bm {\hat e}}_{_2}}_\text{axis} + \underbrace{\theta^{\mathbfcal{J}(j)}\ {\bm {\hat e}}_{_3}}_\text{angle}
	\label{eq:14}
\end{equation}
where:
\begin{itemize}
\setlength{\itemsep}{-0.1cm}
  \item subscript $c$ and superscript $p$ stand for the child and parent frames, respectively, anchored to either a body segment $S$ or an IMU $M$.
  \item ${\bm {\hat e}}_{_i}$ for all $i$ represents a standard-basis vector.
  \item $\phi_p^{\mathbfcal{J}(j)} \in [0,180]$ and $\phi_a^{\mathbfcal{J}(j)} \in [0,360)$ denote the {\itshape polar} and {\itshape azimuthal angles}, respectively, defining a unit axis of rotation in a spherical coordinate system.
  \item $\theta^{\mathbfcal{J}(j)} \in [0,180]$ is angle of rotation about the defined axis.
\end{itemize}

A vectorisation of ${\bm x}$ is performed such that ${\bm x} \in \mathbb{R}^{16\times1} \rightarrow {\bm x} \in \mathbb{R}^{1\times16}$ for convenience of notation. Then, an augmented dictionary variable, ${\bm \Psi}({\bm \tau})$ is defined as the collective pose characterisation vector timeseries obtained through the augmentation of ${\bm \Psi}$

        
        
        
        
         
        

\begin{equation}
	{\bm \Psi}({\bm \tau}) =
	\begin{bmatrix}
        {\bm x}_1(\tau_{_1}) & {\bm x}_2(\tau_{_1}) &  & \cdots & {\bm x}_{12}(\tau_{_1})\\
        
        {\bm x}_1(\tau_{_2}) & {\bm x}_2(\tau_{_2}) &  &  & \\
        
        \vdots &  & \ddots &  & \\
        
         &  &  & {\bm x}_{11}(\tau_{_{N-1}}) & {\bm x}_{12}(\tau_{_{N-1}}) \\
        
        {\bm x}_1(\tau_{_N}) & \cdots &  & {\bm x}_{11}(\tau_{_N}) & {\bm x}_{12}(\tau_{_N})
    \end{bmatrix}
	\label{eq:15}
\end{equation}
where ${\bm \tau} \in \mathbb{Z}^N$ represents the time index vector for the augmented timeseries.

For each arbitrary time index $\tau_{k}$, we sample two Gaussian-distributed noise terms, ${\bm\epsilon_1} \in \mathbb{R}^2$ and $\epsilon_2 \in \mathbb{R}$, to augment the axis-angle representation outlined in \cref{eq:14} as follows
\begin{equation}
    \begin{split}
    	\prescript{p}{}{\bm e}_{c}^{\mathbfcal{J}(j)}(\tau_{k}) &=
    	\begin{bmatrix}
            \phi_p^{\mathbfcal{J}(j)} &
            \phi_a^{\mathbfcal{J}(j)} &
            \theta^{\mathbfcal{J}(j)}
        \end{bmatrix}^T
        +
        \begin{bmatrix}
            {\bm\epsilon_1}\\
            \epsilon_2
        \end{bmatrix}\\
        &=
        (\phi_p^{\mathbfcal{J}(j)}+\delta\phi_p)\ {\bm {\hat e}}_{_1} + (\phi_a^{\mathbfcal{J}(j)}+\delta\phi_a)\ {\bm {\hat e}}_{_2}\\
        &\quad+ (\theta^{\mathbfcal{J}(j)}+\delta\theta)\ {\bm {\hat e}}_{_3}
    \end{split}
	\label{eq:16}
\end{equation}
where ${\bm\epsilon_1}\sim\mathbfcal{N}_1\left(\bm{0}_{2\times1},{\bm \Sigma}\right)$ is used to augment $\phi_p^{\mathbfcal{J}(j)}$ and $\phi_a^{\mathbfcal{J}(j)}$. The {\itshape symmetric covariance matrix}, ${\bm \Sigma}$ is parameterised by a variance $\sigma_\phi^2$
\begin{equation}
	{\bm \Sigma}(\sigma_\phi^2) =
	\begin{bmatrix}
        \sigma_\phi^2 & 0\\
        0 & \sigma_\phi^2
    \end{bmatrix}
	\label{eq:17}
\end{equation}
and $\epsilon_2\sim\mathcal{N}_2(0,\sigma_\theta^2)$ is used to augment $\theta^{\mathbfcal{J}(j)}$ and is parameterised by a variance $\sigma_\theta^2$.

The strength of the proposed data augmentation technique is that it has a single controllable hyperparameter for each of the two main elements defining a static orientation: $\sigma_\phi^2$ and $\sigma_\theta^2$ for the axis and angle of rotation respectively. Assigning different values to both hyperparameters provides varying data augmentation characteristics as per the application requirements. \cref{fig:ouraxes,fig:ourangle} illustrate one possible augmentation result using the proposed method. The carefully noised augmented timeseries resembles the output signals of {\itshape microelectromechanical systems} (MEMS) making up many of today's commercial IMU sensors. Moreover, the addition of noise can be intentionally exaggerated to boost the robustness of trained classifiers.


\subsection{Sleep Posture Classification}  \label{sec:postureclassification}
The definition of the collective pose characterisation vector timeseries is context-dependent; it can be either $\prescript{_{\cross[0.4pt]}}{}{\bm \Psi}(\cdot)$ or $\prescript{*}{}{\bm \Psi}(\cdot)$ to denote training and testing timeseries, respectivey. Herein $(\cdot)$ refers to the time index vector ${\bm t} \in \mathbb{Z}^O$ or ${\bm \tau}$ to indicate real or augmented timeseries respectively.
By definition, a classifier $\mathcal{F} : {\bm x} \rightarrow y$ is required such that $ y \in \mathbfcal{Y} = \{\mathcal{Y}_1,\mathcal{Y}_2,\dots,\mathcal{Y}_{12}\}$ denotes the posture label. For clarity of notation, a generic ${\bm x}$ could either be a training $\prescript{_{\cross[0.4pt]}}{}{\bm x}$ or testing $\prescript{*}{}{\bm x}$, which corresponds to $\prescript{_{\cross[0.4pt]}}{}y$ and $\prescript{*}{}y$ respectively regardless of whether a real or augmented time index is considered.


We leverage on an {\itshape error-correcting output codes} (ECOC) model \cite{Dietterich1995,AllweinEALLWEIN2000} to achieve multi-class classification via aggregating binary classifiers, $f_i : \{i \in \mathbb{Z},\ 1 \leq i \leq L\}$. The ECOC framework begins with the {\itshape encoding step} in which an {\itshape encoding matrix}, $\mathbfcal{M} \in \{-1,0,+1\}^{12 \times L}$, dictates the class memberships for each $f_i$ such that these values denote negative, ignored and positive classes respectively. The matrix elements of $\mathbfcal{M}$ is denoted by $m_j^i$ corresponding to arbitrary class $\mathcal{Y}_j$ and binary classifier $f_i$. Depending on the adopted encoding strategy, the number of employed binary classifiers and their collective generalisation capability may vary. Herein, we use the {\itshape one-against-one} encoding technique that explores all possible pairs of classes $(L=66)$, as this was found to have good generalisation capability, without compromising computational efficiency \cite{Hsu2002}.

Once all binary classifiers are fully trained, the ECOC model then relies on a {\itshape decoding step} to map the output of $f_i$ to the corresponding class label. To accomplish this, a base of reference codewords is created to define the aggregate outputs from all classifiers for each class. The ECOC model eventually compares a given test codeword against each of the reference codewords to determine the class of the largest likelihood.  Different decoding strategies were proposed in the literature with the most popular ones being (1) {\itshape distance-based}, (2) {\itshape probabilistic-based} and (3) {\itshape pattern space transformation} techniques \cite{Escalera2010}. In this work, the pairwise {\itshape Hamming distance} is used as the loss measure to estimate the most likely class label $\prescript{*}{}{\hat y}_j(t_k)$ for $\prescript{*}{}{\bm x}_j(t_k)$, i.e.
\begin{equation}
	\prescript{*}{}{\hat y}_j(t_k) = \argmin_{\mathcal{Y}_j \in {\bm Y}} \frac{1}{2L} \ \sum_{i} 1-\text{\itshape sgn}\left(m_j^i \cdot f_i\left(\prescript{*}{}{\bm x}_j(t_k)\right)\right)
	\label{eq:18}
\end{equation}
where $m_j^i \in \{+1,-1\}\ \forall i\forall j$. In this context, $m_j^i$ serves as the groundtruth binary label for $f_i(\cdot)$ given $\mathcal{Y}_j$.
A similar expression can be formulated for $\prescript{*}{}{\hat y}_j(\tau_k)$ and $\prescript{*}{}{\bm x}_j(\tau_k)$ too.

In regard to the binary classifiers, we apply an ensemble of {\itshape soft margin} SVM algorithms to $\prescript{_{\cross[0.4pt]}}{}{\bm \Psi}({\bm \tau})$ obtained from the one-shot learning phase. The framework of this algorithm uses two hyperparameters: (1) a slack variable $\xi_{k,j}$ to tolerate minimal misclassifications owing to outliers in the training dataset, and (2) a scalar $C$ to control the smoothness of the classifier's decision boundary. The standard optimisation problem for each $f_i$ is outlined in \cref{eq:19}, where only one positive and one negative classes are selected according to the one-against-one encoding defined by $\mathbfcal{M}$. While searching for a solution hyperplane, parameterised by weight vector ${\bm W} \in \mathbb{R}^{1\times2N}\ \text{and a scalar bias}\ b$, a {\itshape Gaussian kernel} ${\bm \varphi}_\gamma : \mathbb{R}^{1\times16} \rightarrow \mathbb{R}^{1\times2N}$ of hyperparameter spread $\gamma$ is applied to the {\itshape support vectors} to enhance the separability of classes \cite{Abe2010}. Finally, a Bayesian optimisation algorithm \cite{Shahriari2016} is used to find the optimal values of the aforementioned hyperparameters:
\begin{equation}
	\min_{{\bm W},b} \quad \frac{1}{2} \ {\norm{\bm W}}^2 + C \ \sum_{k}\sum_{j}{\xi_{k,j}}\\
	\label{eq:19}
\end{equation}
\begin{equation*}
    \begin{aligned}
		\textrm{s.t.} \quad & \! m_j^i \cdot \left({\bm \varphi}_\gamma\left(\prescript{_{\cross[0.4pt]}}{}{\bm x}_j(\tau_k)\right) \cdot {\bm W}^T + b \right) \geqslant 1-\xi_{k,j}\\
		&\xi_{k,j} \geqslant 0\\
		&i : f_i \in \mathcal{F}\\
		&j : j \in \mathbb{Z},\ \forall m_j^i \in \{+1,-1\}\\
		&k : k \in \mathbb{Z},\ 1 \leq k \leq N
	\end{aligned}
\end{equation*}

\section{Experimental Setup}  \label{sec:experiments}
This section describes the experimental design and setup for implementing the posture learning framework reported in \cref{sec:methods}, for both the virtual and the human participant pipelines from data collection all through performance evaluation and interpretation.

\begin{figure*}[!b]
    \centering
    \includegraphics[width=0.8\textwidth]{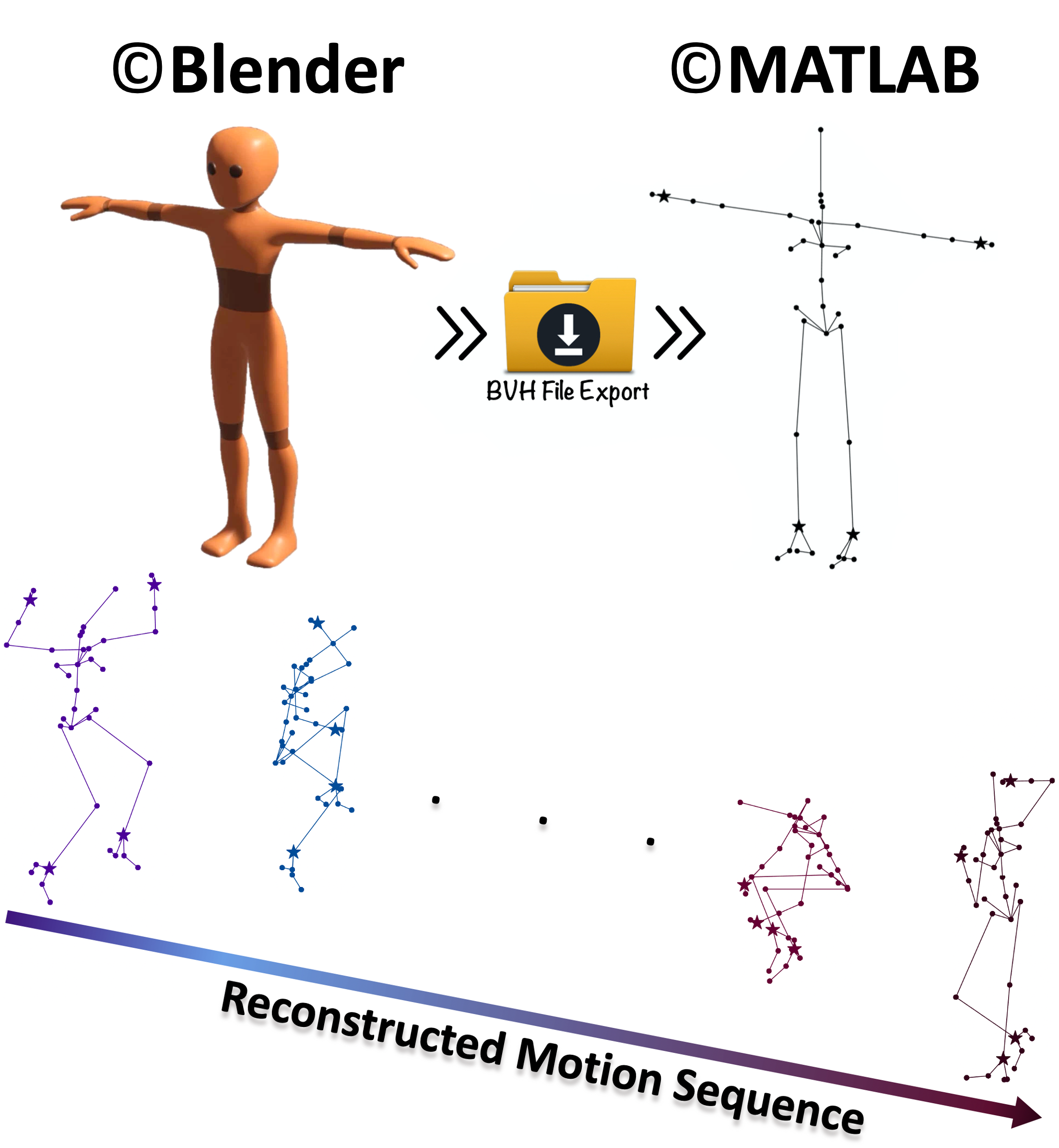}
    \caption{Reconstruction of {\itshape in silico} sleep motion sequence in MATLAB$^\copyright$. Pentagram symbols are used to annotate the four extremity limb distal joints defining ${\bm \rchi}^v$ at each sleep posture.}
    \label{fig:PosesReconstruction}
\end{figure*}

\subsection{Virtual Sleep Pipeline}  \label{sec:pipelinevirtualsleep}
As mentioned in \cref{sec:virtualsleep}, {\itshape in silico} sleep simulation is built around a motion sequence animated in Blender$^\copyright$ through manually keyframing each sleeping pose as depicted in \cref{fig:insilicoexpa}. The motion sequence keeps each pose for ten consecutive frames before making another ten-frame transition to the next pose, thus making the whole animation 230 frames long in total. The animation relies on linear interpolation to fill in the gaps between each two consecutive keyframes.

The motion sequence is then exported from Blender$^\copyright$ in the BVH file format and imported into the MATLAB$^\copyright$ (The MathWorks, Massachusetts, US) environment via a bespoke parser script. The parser creates a data structure to allow for the reconstruction of $\bm{B}$ throughout the motion sequence as shown in \cref{fig:PosesReconstruction}. Another pose characterisation script then extracts ${\bm \rchi}^v$ at each keyframed sleep posture, forming the pose characterisation vector ${\bm \Psi}^v$ to be used in the one-shot learning scheme explained in \cref{sec:oneshotlearning}.

\subsection{Participant Study Pipeline}  \label{sec:pipelineparticipantstudy}
An experimental setup was built at an outdoor university facility for ideal data collection conditions, avoiding measurement anomalies due to, for example, interference from the building environment with the magnetometer. Prior to the pilot experiment, all IMU sensors were calibrated as described in \cite{Woodman2007,Kok2017} to estimate and reduce errors owing to constant bias, scale factors, cross-axis sensitivity and response nonlinearity. The protocol was approved  by The University of Liverpool Research Ethics Committee (review reference: 9850).

The microcontroller chip built in each wearable sensor performs uniform sampling of both IMUs at a rate of $30$ Hz. For optimal multi-sensor data transmission, dual-IMU data packets are simultaneously sent over Bluetooth from all four wearable sensor modules (clients) to the localhost server running a Python script. All data packets are timestamped using a monotonic digital clock with a microsecond resolution. At the end of the data collection session and after sensor fusion, these timestamps are later needed to synchronise, using linear interpolation, quad-sensor relative orientations under one unified time vector as illustrated in \cref{fig:systemfigure}.

\begin{figure}[t]
    \centering
    \includegraphics[width=0.7\columnwidth]{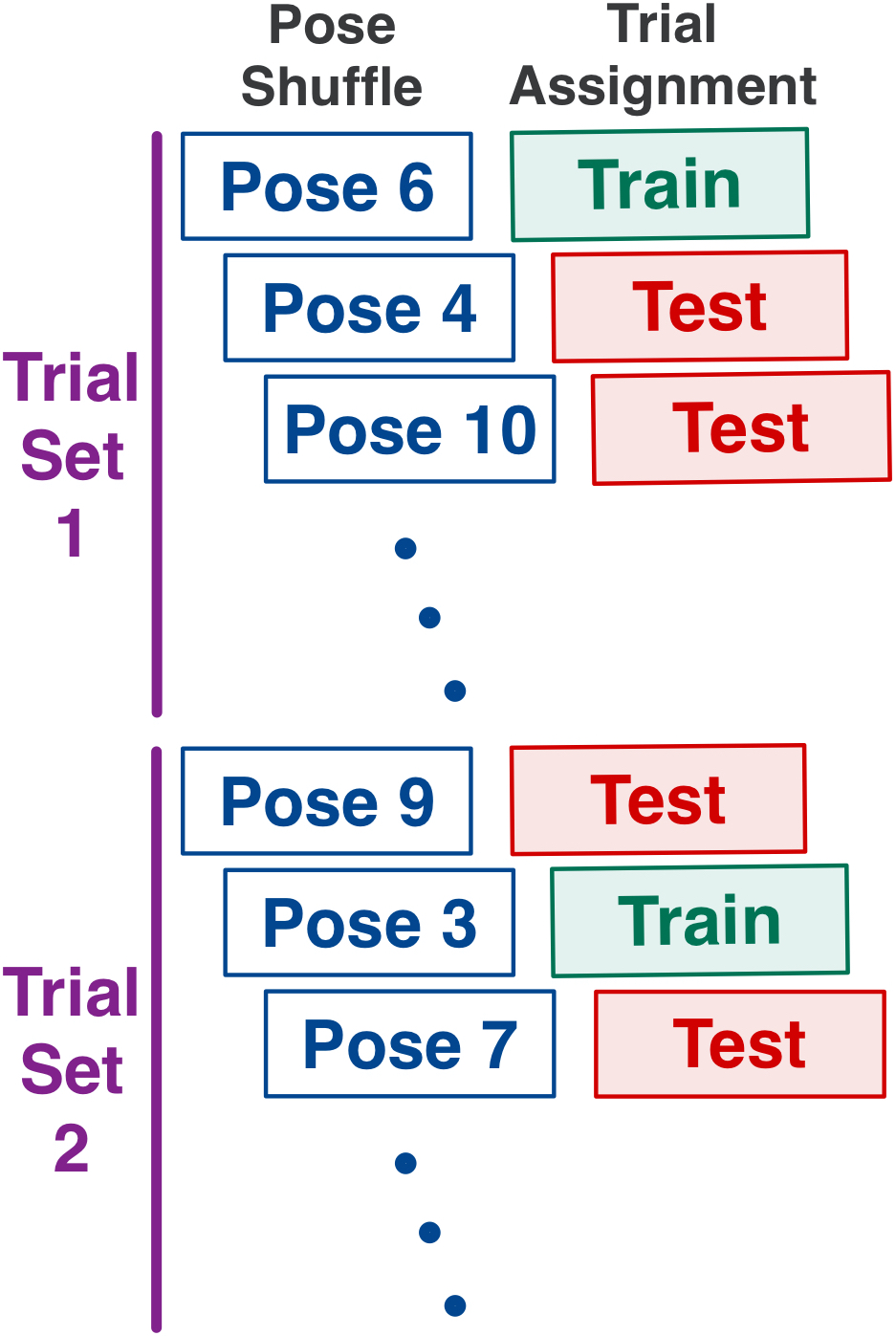}
    \caption{Visual illustration of the data collection protocol for the human participant experiment.}
    \label{fig:DataCollectionProtocol}
\end{figure}

Both IMU sensors were placed on each extremity limb such that they are approximately aligned with the distal joint axes to gauge nearly accurate segment-to-segment orientations. As depicted in \cref{fig:handsensors}, the $y$-axis and $z$-axis of both IMUs were aligned as much as possible with the flexion/extension and ulnar/radial deviation axes of the wrist joint when the hand is parallel to the forearm. Both IMUs were positioned maximally close towards the wrist joint to reduce artefacts from muscle contractions and skin movements, and to avoid interference with the elbow rotation. Similar considerations were taken into account for the placement of the lower limb sensor modules.

A leaflet containing pictures of the sleep postures was given to the participant to assist them in replicating the desired poses before each sample was recorded. As portrayed in \cref{fig:DataCollectionProtocol}, each sleep posture is recorded twice; one recording for each of two trial sets. To ensure postural data resembles that of a realistic sleep scenario, we collect statistically independent posture samples using a random pose shuffling technique throughout each trial set. In addition, to account for the participant gaining familiarity with pose replication over the course of the experiment, we adopt a randomised train/test trial assignment strategy.

At the server back end, all received sensor data are immediately logged into a {\itshape comma-separated values} (CSV) file for subsequent import into MATLAB$^\copyright$. Therein, a sensor fusion script applies Madgwick filtering to each IMU data channels to estimate its orientation given a unit quaternion as the initial orientation estimate and learning rate $\beta = 0.1$. All IMU orientations are then collectively fed into a pose characterisation script which extracts ${\bm \rchi}^w$ belonging to each train-labelled posture recording, and $\prescript{*}{}{\bm \Psi}^w({\bm t})$ from test-labelled trials. For each training posture trial, only one randomly selected sample from the quad-sensor relative orientation timeseries is used to identify ${\bm \rchi}^w$ for that pose. The resultant ${\bm \Psi}^w$ will be later utilised in the one-shot learning described in \cref{sec:oneshotlearning}. When constructing the $\prescript{*}{}{\bm \Psi}^w({\bm t})$ timeseries, the time vector ${\bm t}$ has a length $O = max(\Omega_j)$, where $\Omega_j$ is the timeseries length of any $j^{\text{th}}$ test-labelled posture recording. Thus, $\prescript{*}{}{\bm \Psi}^w({\bm t})$ can be mathematically written as follows:

\begin{equation}
    \prescript{*}{}{\bm \Psi}^w({\bm t}) =
    \left[
    \begin{array}{c c c}
        \prescript{*}{}{\bm x}_1({\bm t}_{1:\Omega_1}) & \cdots & \prescript{*}{}{\bm x}_{12}({\bm t}_{1:\Omega_{12}})\\
         & & \\
        {\bm \Phi}(\bar{\bm t}_{1:\Omega_1}) & \cdots & {\bm \Phi}(\bar{\bm t}_{1:\Omega_{12}})
    \end{array}
    \right]
\label{eq:20}
\end{equation}
such that $\bar{\bm t}_{1:\Omega_j}$ is the relative complement of ${\bm t}_{1:\Omega_j}$ in ${\bm t}$:
\begin{equation*}
	\bar{\bm t}_{1:\Omega_j}
	= {\bm t} \setminus {\bm t}_{1:\Omega_j}
	\label{eq:20_1}
\end{equation*}
and ${\bm \Phi}(\bar{\bm t}_{1:\Omega_j})$ denotes a sixteen-column matrix of {\itshape Not a Number} (NaN) elements to account for any $j^{\text{th}}$ test-labelled posture recording with $\Omega_j < O$.

\subsection{One-Shot Learning}  \label{sec:oneshotlearning}
In scenarios where only scarce data is available, data augmentation is necessary. As outlined in \cref{sec:dataaugmentation}, we proposed a one-shot learning method for modelling human sleep postures given a single observation per pose. Depending on whether the virtual or human participant pipeline is considered, the usage of data augmentation varied slightly.

For the {\itshape in silico} sleep simulation, the motion sequence only provides ${\bm \Psi}^v$, meaning that separate training and testing timeseries are unavailable for ML. In such case, data augmentation is employed twice for generating training and testing timeseries, $\prescript{_{\cross[0.4pt]}}{}{\bm \Psi}^v({\bm \tau})$ and $\prescript{*}{}{\bm \Psi}^v({\bm \tau})$ respectively. Besides the single posture observation, new augmented samples $(N = 499)$ are appended to the $\prescript{_{\cross[0.4pt]}}{}{\bm \Psi}^v({\bm \tau})$ timeseries, contributing to a total of $500$ training samples per sleep posture. Additional augmented samples $(N = 125)$ are designated for $\prescript{*}{}{\bm \Psi}^v({\bm \tau})$.


With regard to the human participant experiment, since $\prescript{*}{}{\bm \Psi}^w({\bm t})$ is obtained from the test-labelled trial recordings, it is therefore required to generate only one timeseries for classifier training that is $\prescript{_{\cross[0.4pt]}}{}{\bm \Psi}^w({\bm \tau})$. Hence, for each posture, augmented samples $(N = 999)$ are appended to the single observation, contributing to a total of $1000$ training samples per sleep posture.

The timeseries augmentation step described in \cref{sec:dataaugmentation} for both the virtual and human participant experiments was carried out given the same range of hyperparameter settings, $(\sigma_\phi^2, \sigma_\theta^2)$. A grid ${\bm \Re} \in \mathbb{Z}^{6\times6}$ of discrete points in the hyperparameter space was constructed where $\sigma_\phi^2 \in {\bm \Re}_\phi = \{20, 200, 400, 600, 800, 1000\}$ and $\sigma_\theta^2 \in {\bm \Re}_\theta = \{20, 100, 200, 300, 400, 500\}$, yielding 36 different data augmentation settings. Given each pair of hyperparameters, the respective training and testing timeseries datasets are used for the training and testing of the posture classification algorithm outlined in \cref{sec:postureclassification}. For the soft margin SVM problem, the Bayesian optimisation algorithm carries out iterative search ($60$ iterations) for the optimal values of the two hyperparameters $C$ and $\gamma$ over the range $\left[1e^{-3},1e^3\right]$. 


\subsection{Performance Evaluation}  \label{sec:PerformanceEvaluation}
Two main metrics are used for the evaluation of the posture classification performance; the {\itshape accuracy} $(m_{acc})$ and {\itshape F1 score} $(m_{F1})$. The accuracy refers to the ratio of correct classifications to the total number of testing samples, and is reliable when the testing dataset is evenly distributed as it is the case with the virtual sleep experiment. For class-imbalanced datasets, such as $\prescript{*}{}{\bm \Psi}^w({\bm t})$, the F1 score offers a less biased assessment of the model performance through finding the harmonic mean of {\itshape precision} and {\itshape recall} \cite{He2013}. To mitigate any skewed dataset distribution, we employ a macro-averaged F1 score expressed in \cref{eq:21} that involves computing all class-specific scores independently, followed by finding the overall unweighted arithmetic mean.

\begin{equation}
	m_{F1} = \frac{1}{12} \sum_{k=1}^{12} \left(2 \times \frac{\text{recall}(j) \times \text{precision}(j)}{\text{recall}(j) + \text{precision}(j)}\right)
	\label{eq:21}
\end{equation}
such that
\begin{equation*}
    \begin{split}
	\text{recall}(j)
	&= \frac{\text{TP}(j)}{\text{TP}(j) + \text{FN}(j)}\\
	\text{precision}(j)
	&= \frac{\text{TP}(j)}{\text{TP}(j) + \text{FP}(j)}
	\end{split}
	\label{eq:21_1}
\end{equation*}
and $\text{TP}(j)$, $\text{FP}(j)$ and $\text{FN}(j)$ correspond to the true positives, false positives and false negatives, respectively, of a given arbitrary $j^\text{th}$ posture class.

Additionally, all performance evaluation experiments are repeated ten times where the mean and standard deviation of both metrics indicate the effectiveness of the Bayesian optimisation algorithm in solving for the SVM optimal hyperparameters.

\subsection{Performance Interpretation}  \label{sec:PerformanceInterpretation}
The metrics described in \cref{sec:PerformanceEvaluation} are used to monitor, measure and compare the performance of one or more models during the training and testing phases. To build confidence in deploying ML algorithms in real world applications, additional interpretation methods need to be created to allow human users (e.g. clinicians) to comprehend and trust the outputs and decisions made available by these algorithms. Ideally, these methods are expected to unravel the reasoning behind ML algorithms, and be able to explain their cases of success and failure.

Therefore, we herein present two approaches to lend more explainability to the posture learning algorithm. These approaches are employed to explore any interesting data trends, and the findings are then used to interpret the model's posture inference.

The first visualisation-based approach utilises {\itshape uniform manifold
approximation and projection} (UMAP) \cite{McInnes2018} to produce a two-dimensional (2D) {\itshape force-directed graph} $\mathbfcal{U}$ of high-dimensional datasets, such as $\prescript{_{\cross[0.4pt]}}{}{\bm \Psi}({\bm \tau})$ and $\prescript{*}{}{\bm \Psi}(\cdot)$. Dimensionality reduction has been successfully applied to visualise data in many domains, including human motion analysis \cite{Airaksinen2020}, pedagogical research \cite{Elnaggar2021} and speaker recognition \cite{Elnaggar2019b}. In this work, UMAP facilitates the visualisation of postural observations of the same posture (intra-class distribution) as well as across different postures (inter-class distribution). Such data analysis could provide insights on the research methods (e.g. assessing the added value of data augmentation), and on the problem as a whole (e.g. estimating human postural variability). The visualisation software was based on a MATLAB implementation of UMAP \cite{Meehan2022}.

Although UMAP is known for its capability to preserve the local and global data structures in $\mathbfcal{U}$, it lacks a guarantee on the faithful reconstruction of the actual cluster sizes and inter-cluster distances. Therefore, further interpretation tools, perhaps metric-based, are required for a fine-resolution analysis.


\begin{figure*}[!b]
    \centering
    \begin{subfigure}[b]{0.49\textwidth}
        \centering
    	\includegraphics[width=\linewidth]{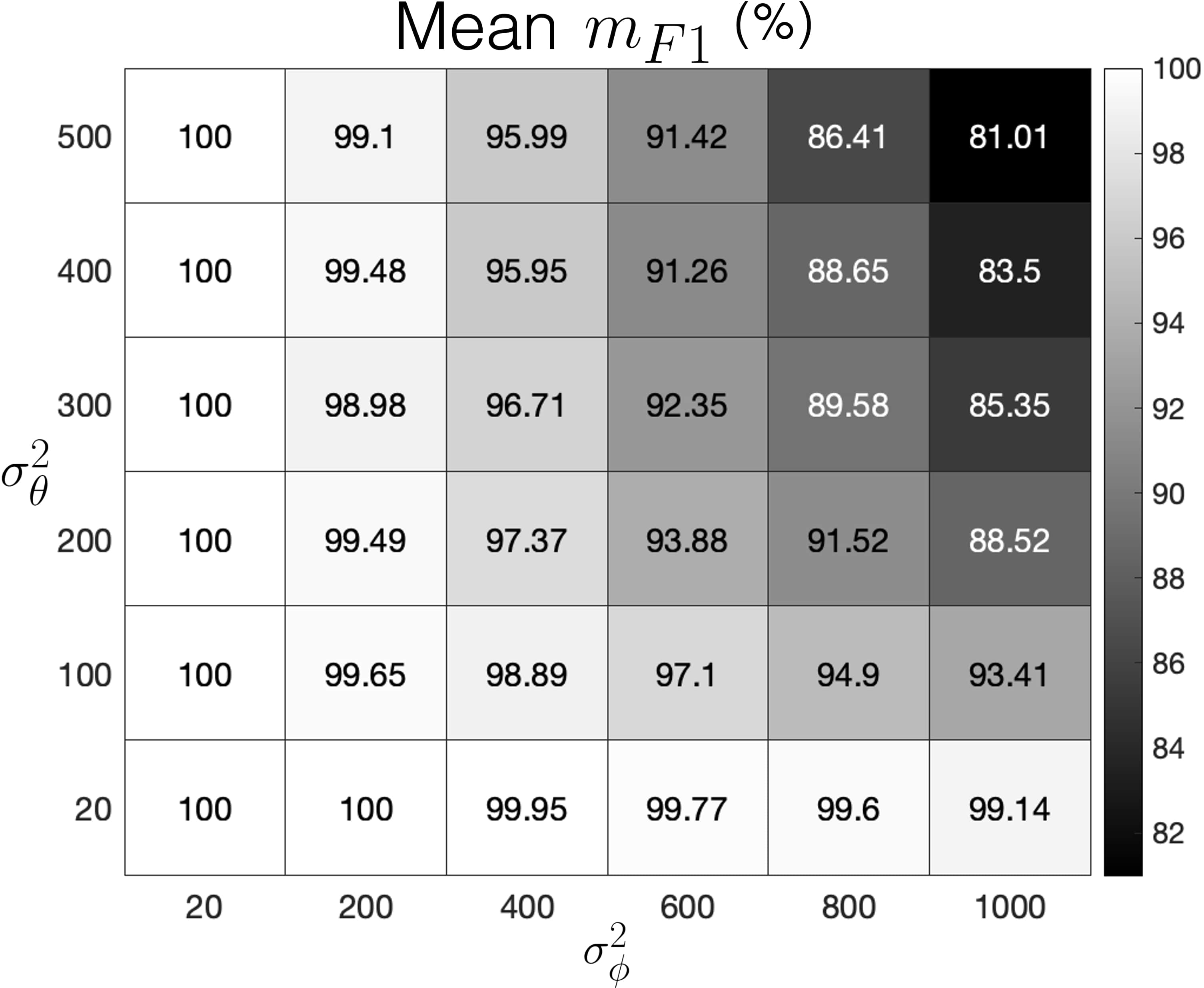}
    	\caption[Optional Caption]{}
    	\label{fig:virtualmeanF1}
    \end{subfigure}
    \begin{subfigure}[b]{0.49\textwidth}
        \centering
    	\includegraphics[width=\linewidth]{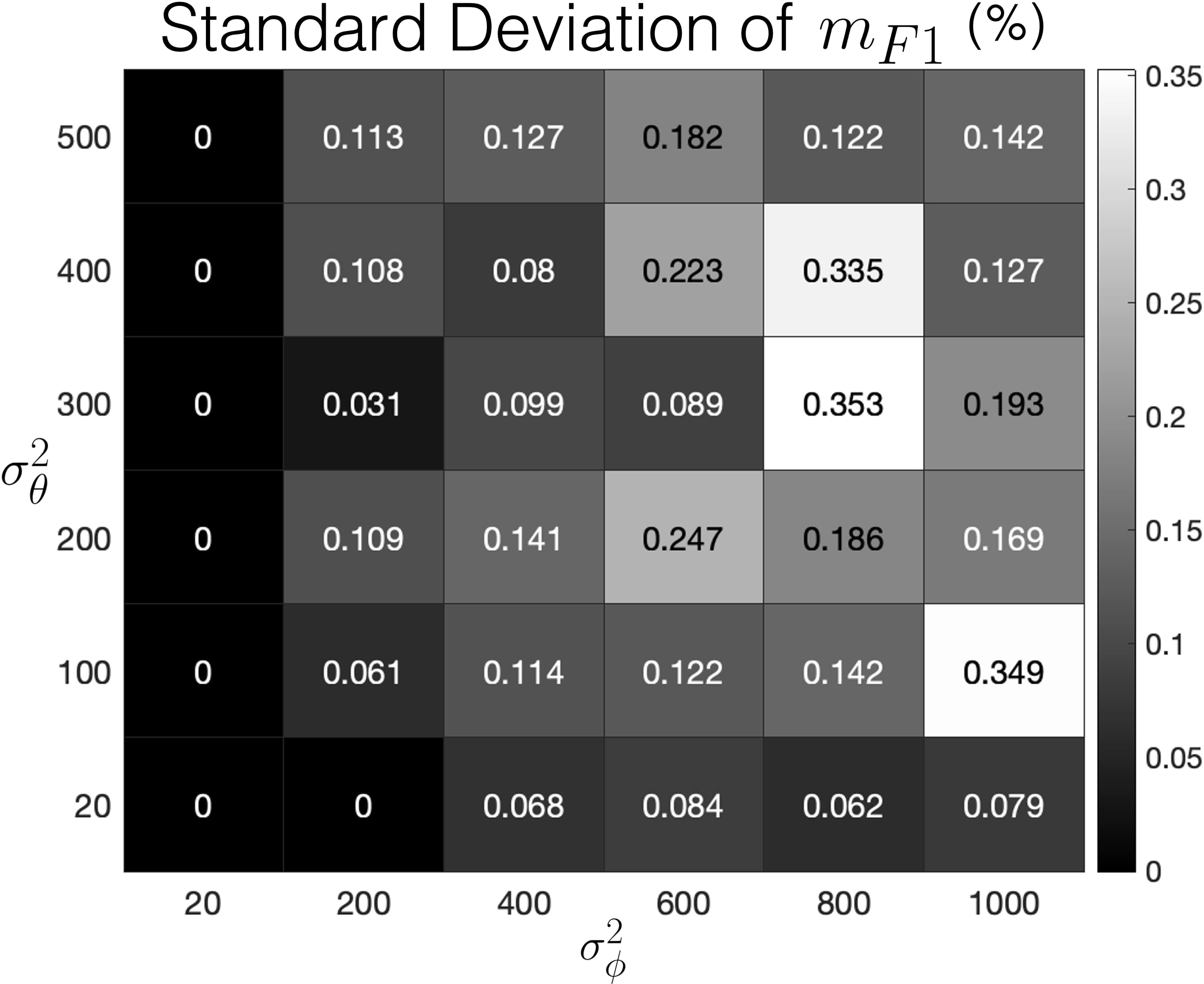}
    	\caption[Optional Caption]{}
    	\label{fig:virtualstdF1}
    \end{subfigure}
    \caption{Performance evaluation metrics for the virtual sleep experiment.}
    \label{fig:virtualmeanandstdF1}
\end{figure*}

Localisation and pose estimation approaches often use quaternions for tracking the orientation of some target asset. Several works reported the use of the angular offset $\Delta \theta_{a,b}$, expressed in \cref{eq:22}, between any two quaternions ${\bm q}_a$ and ${\bm q}_b$ as a common metric to assess the (dis)similarity between orientations \cite{Taetz2016,Sola2017,Kraft2003}:
\begin{equation}
	\Delta \theta_{a,b} = 2 \arccos{\left[{\bm q}_{a,b}^\epsilon\right]_w}
	\label{eq:22}
\end{equation}
where ${\bm q}_{a,b}^\epsilon = {\bm q}_a^* \otimes {\bm q}_b$ represents the residual orientation error between ${\bm q}_a$ and ${\bm q}_b$, and $\left[\ \cdot\ \right]_w$ is an operation to extract the scalar term of ${\bm q}_{a,b}^\epsilon$.

Such error metric can be used to fuse different orientation estimates into one more robust estimate, or compare different estimation techniques. Although it is useful for some applications, it completely overlooks the axis of rotation error in ${\bm q}_{a,b}^\epsilon$. For the purpose of this paper it is essential to identify where any postural discrepancies/overlaps, may emerge from: {\itshape are they due to the angle, axis or both components of the extremity limb orientations?} In fact, such information can be used to interpret the model perception, and potentially enable clinicians to make evidence-backed future changes to the sleep analysis problem itself, such as insertion/removal of postures, or altering the pose characterisation method.

Therefore, we propose a second performance interpretation approach empowered by a hybrid metric $\Lambda$ to evaluate the (dis)similarity between multiple posture observations, fusing the axes similarity $\Lambda_\phi$ with the angles similarity $\Lambda_\theta$. Suppose we have two arbitrary postural observations ${\bm x}_a$ and ${\bm x}_b$, then $\Lambda$ can be defined as
\begin{equation}
	\Lambda = \Lambda_\phi + \Lambda_\theta
	\label{eq:23}
\end{equation}
where
\begin{equation*}
    \begin{split}
        \Lambda_\phi &= \sum_{j=1}^{4} \left({\bm x}_{a,1:3}^{\mathbfcal{J}(j)} \cdot {\bm x}_{b,1:3}^{\mathbfcal{J}(j)}\right)\\
        \Lambda_\theta &= \frac{4\pi-\sum_{j=1}^{4} \abs{{\bm x}_{a,4}^{\mathbfcal{J}(j)} - {\bm x}_{b,4}^{\mathbfcal{J}(j)}}}{\pi}
    \end{split}
\end{equation*}
The quadruple axes similarity is captured in $\Lambda_\phi$ using the vector dot product of ${\bm x}_{a,1:3}^{\mathbfcal{J}(j)}$ and ${\bm x}_{b,1:3}^{\mathbfcal{J}(j)}$, whereas $\Lambda_\theta$ computes a normalised similarity measure based on the total absolute angle error between ${\bm x}_{a,4}^{\mathbfcal{J}(j)}$ and ${\bm x}_{b,4}^{\mathbfcal{J}(j)}$. Each of $\Lambda_\phi$ and $\Lambda_\theta$ is defined $\forall j \in \left[1,4\right]$ and scored out of four (i.e. a full score dictates ${\bm x}_a = {\bm x}_b$), contributing to a total similarity score out of eight for $\Lambda$.

\begin{figure*}[!b]
    \centering
    \begin{subfigure}[b]{0.49\textwidth}
        \centering
    	\includegraphics[width=\linewidth]{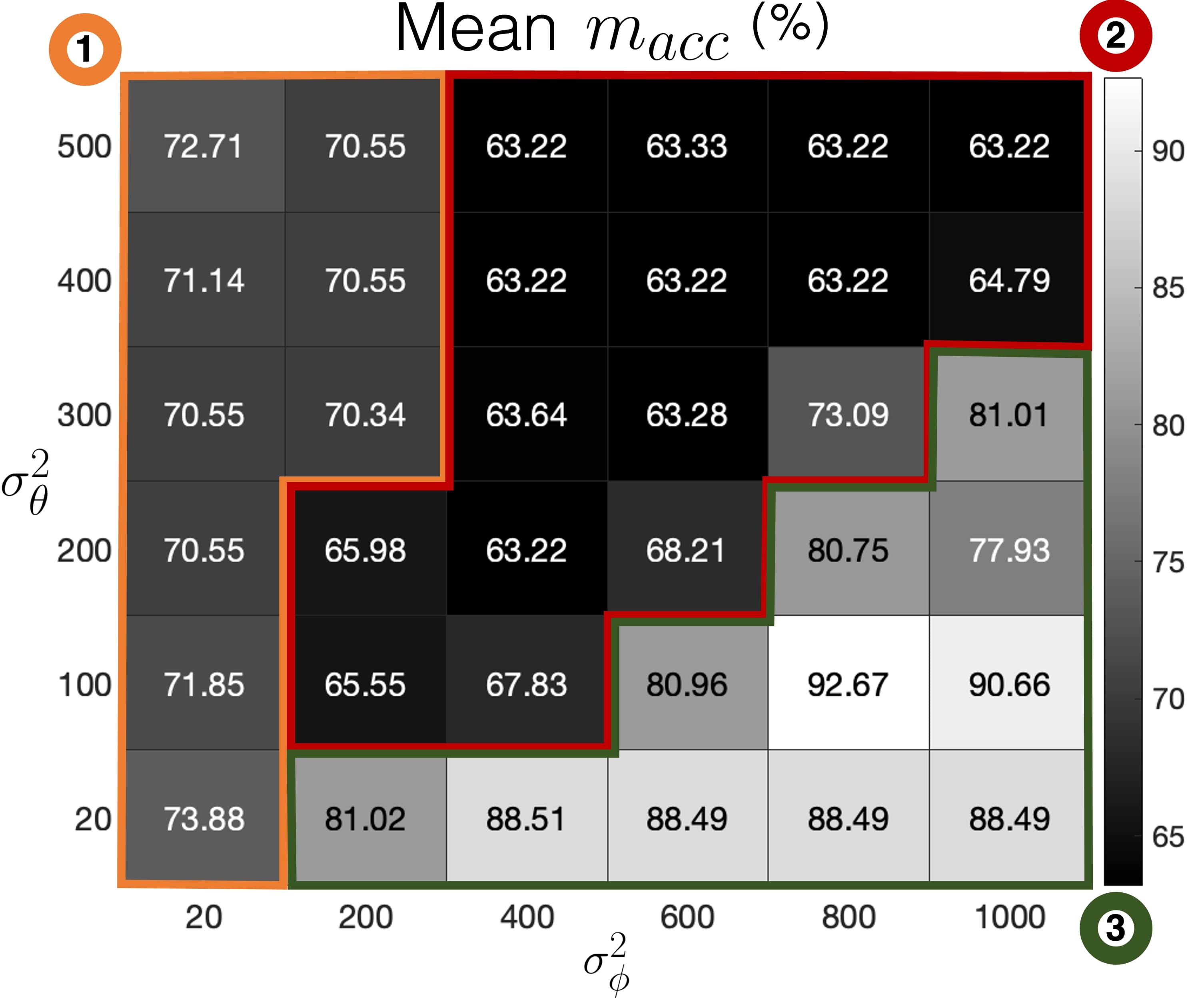}
    	\caption[Optional Caption]{}
    	\label{fig:realmeanacc}
    \end{subfigure}
    \begin{subfigure}[b]{0.49\textwidth}
        \centering
    	\includegraphics[width=\linewidth]{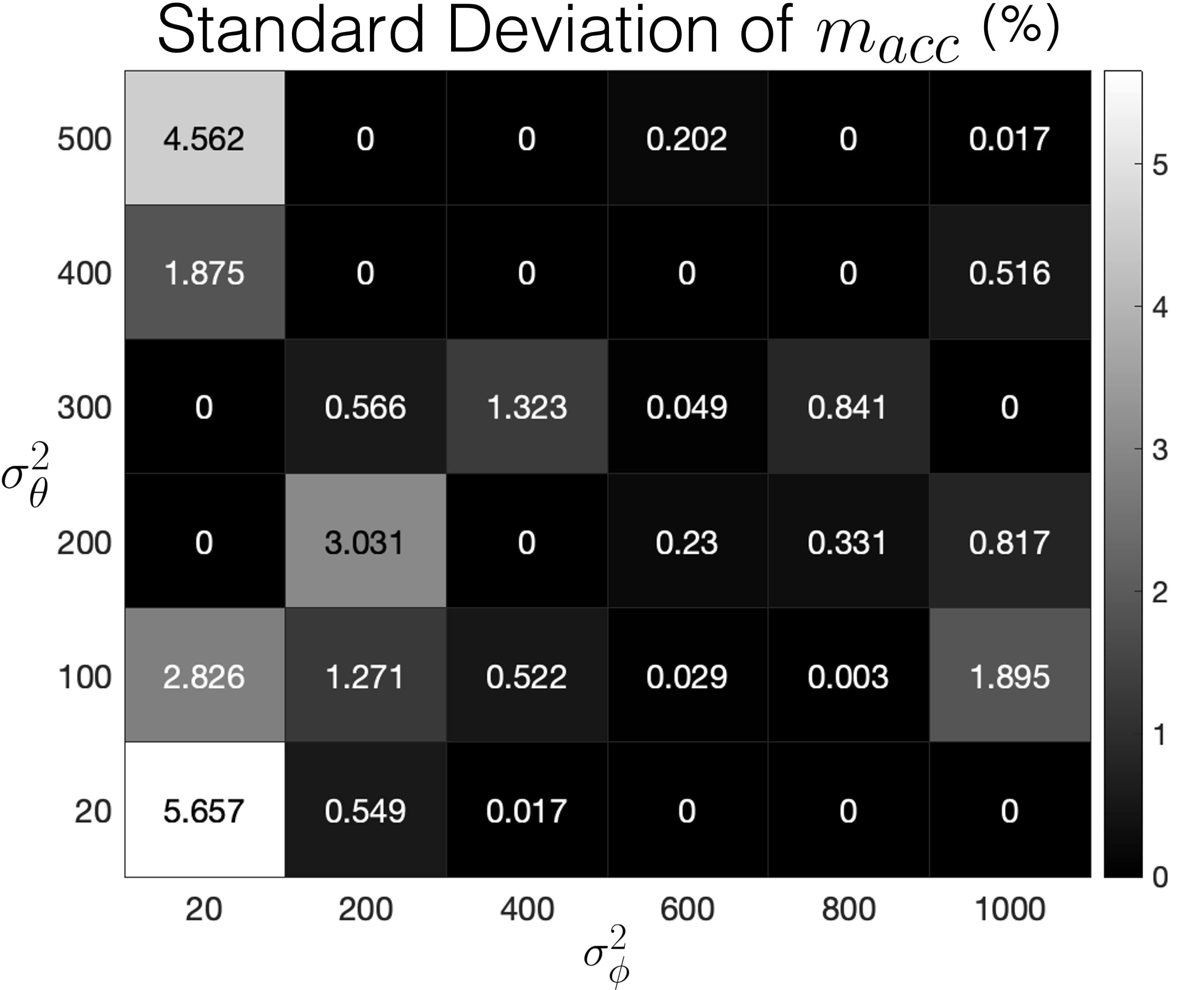}
    	\caption[Optional Caption]{}
    	\label{fig:realstdacc}
    \end{subfigure}
    \par
    \begin{subfigure}[b]{0.49\textwidth}
        \centering
    	\includegraphics[width=\linewidth]{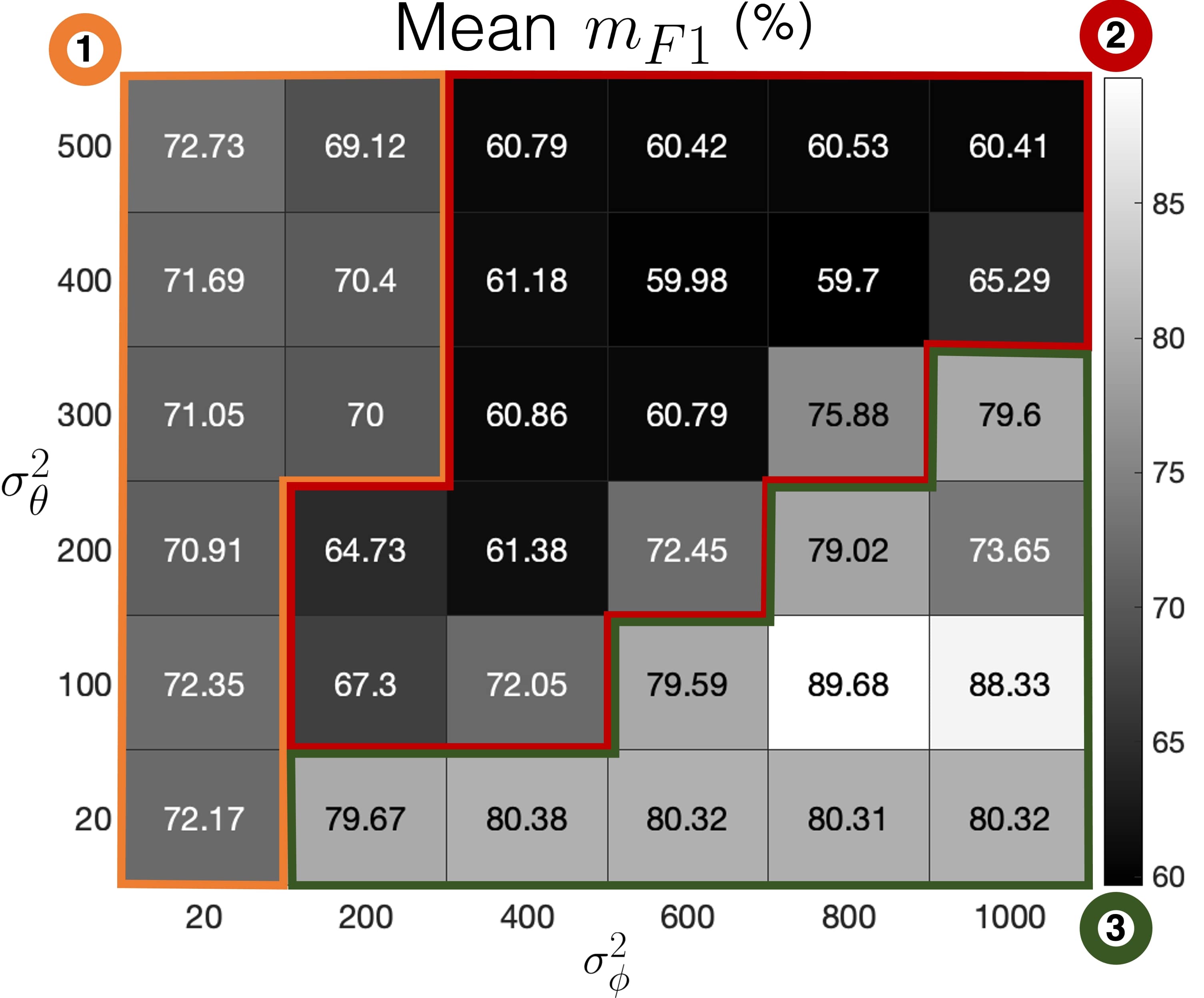}
    	\caption[Optional Caption]{}
    	\label{fig:realmeanF1}
    \end{subfigure}
    \begin{subfigure}[b]{0.49\textwidth}
        \centering
    	\includegraphics[width=\linewidth]{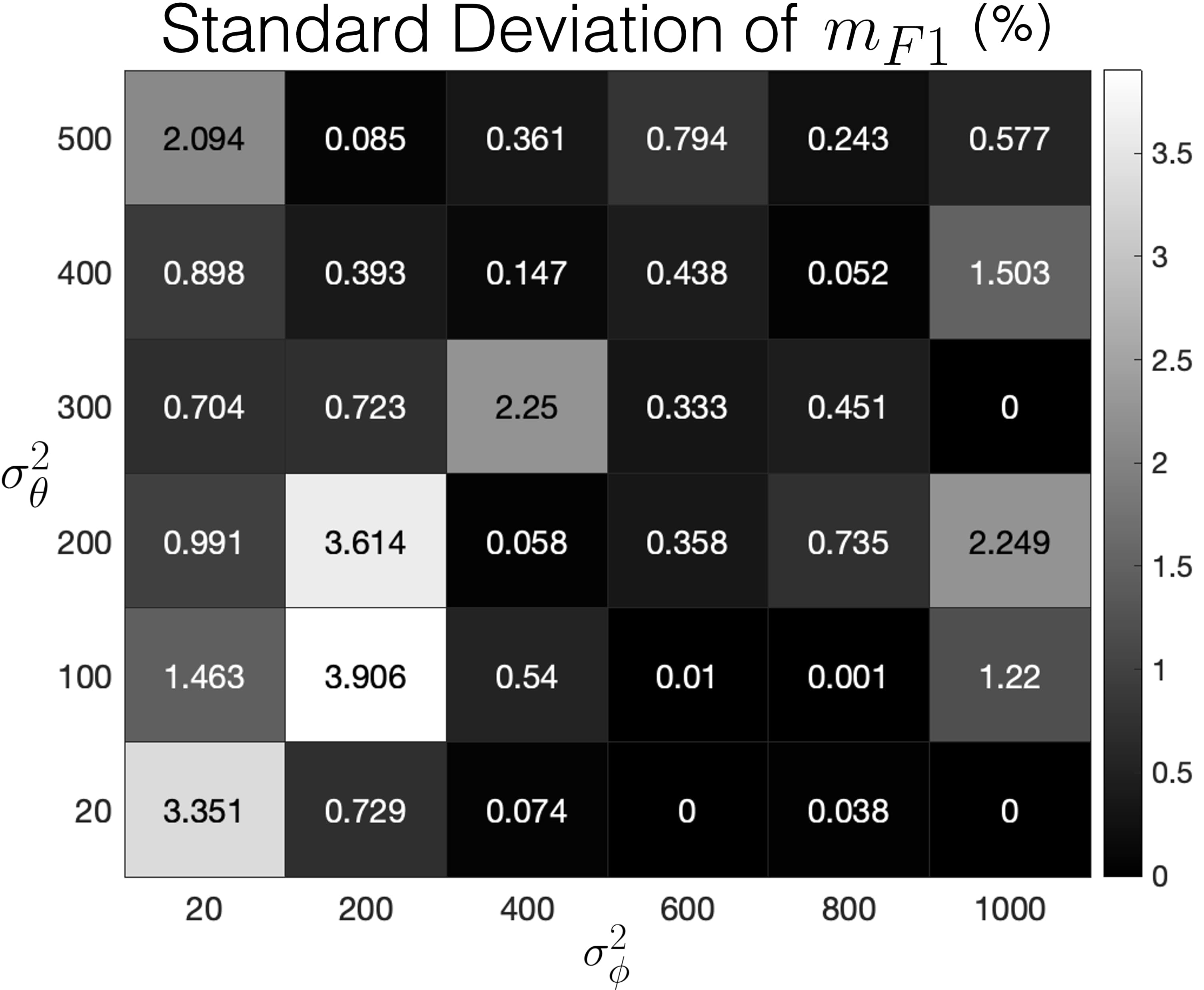}
    	\caption[Optional Caption]{}
    	\label{fig:realstdF1}
    \end{subfigure}
    \caption{Performance evaluation metrics for the human participant pilot experiment.}
    \label{fig:realmeanandstdaccuracyandF1}
\end{figure*}

\section{Results and Discussion}  \label{sec:resultsdiscussion}
This section presents the results obtained following the proposed experimental methods and protocols covered in \cref{sec:methods,sec:experiments}. The discussion first sheds light on the role data augmentation plays in both the {\itshape in silico} (\cref{sec:resultsvirtualexperiment}) and the human participant posture analysis pipelines (\cref{sec:resultsparticipantexperiment}). Thereafter, further performance interpretation uncovers qualitative and quantitative insights on sleep postures and the classification problem as a whole. A comparison of the results obtained with the proposed approach and the state-of-the-art available in the literature is reported afterwards in \cref{sec:standingandfuturedirections}.

\subsection{Virtual Sleep Experiment}  \label{sec:resultsvirtualexperiment}
The {\itshape in silico} sleep posture learning pipeline operates on augmented posture datasets in both the training and testing phases.
The same data augmentation hyperparameters ($\sigma_\phi^2$ and $\sigma_\theta^2$) are shared by $\prescript{_{\cross[0.4pt]}}{}{\bm \Psi}^v({\bm \tau})$ and $\prescript{*}{}{\bm \Psi}^v({\bm \tau})$, thus the posture classification model is tested on postural observations for which the level of variability is known {\itshape a priori}.
As presented in \cref{sec:oneshotlearning}, this evaluation is conducted repeatedly at different levels of postural variability as dictated by the hyperparameter settings, $\sigma_\phi^2$ and $\sigma_\theta^2$, in ${\bm \Re}$. Therefore, the results obtained from the {\itshape in silico} pipeline inform on how sensitive the posture learning framework is to variations in postural observations.


\cref{fig:virtualmeanandstdF1} shows the sleep posture classification performance given each augmentation setting in ${\bm \Re}$. Since all posture classes are equally weighted in $\prescript{_{\cross[0.4pt]}}{}{\bm \Psi}^v({\bm \tau})$ and $\prescript{*}{}{\bm \Psi}^v({\bm \tau})$, we only show the mean and standard deviation of $m_{F1}$ as they are almost identical to that of $m_{acc}$.
At the bottom left corner of ${\bm \Re}$ where the level of injected noise is lowest, $(\sigma_\phi^2,\sigma_\theta^2) = (20,20)$, a perfect $100\%$ $m_{F1}$ score  with zero standard deviation is attained. On the opposite corner, where $(\sigma_\phi^2,\sigma_\theta^2) = (1000,500)$, exaggerated noise injection seems to pose more challenge to the classification task; nevertheless, the mean $m_{F1}$ remains above $81\%$ with small standard deviation.

The mean $m_{F1}$ heat map can be used to study the effect of each augmentation hyperparameter. When $\sigma_\phi^2$ is below $200$, the gradual increase of $\sigma_\theta^2$ barely affects the classification performance. On the other hand, the increase in $\sigma_\phi^2$ tend to have more influence over the performance when $\sigma_\theta^2$ is below $100$. As we move diagonally along ${\bm \Re}$ and near to the top right corner, the influence of stepping up $\sigma_\theta^2$ overtakes that of $\sigma_\phi^2$. Overall, the results obtained through the virtual experiment suggest that the proposed sleep posture learning framework is robust to mild-to-extreme variations in postural observations.

\subsection{Participant Pilot Experiment}  \label{sec:resultsparticipantexperiment}
The participant study pipeline utilises one-shot learning only during the training phase, then validates the resultant trained model on ``unseen'' real timeseries. Therefore, some level of discrepancy (mismatch) is present between the augmented training dataset and test-labelled posture recordings.
Therefore, the participant study complements the virtual experiment through exploring what data augmentation offers to sleep posture learning when unknown discrepancy exists between $\prescript{_{\cross[0.4pt]}}{}{\bm \Psi}^w({\bm \tau})$ and $\prescript{*}{}{\bm \Psi}^w({\bm t})$.



\begin{figure*}[!b]
    \centering
    \begin{subfigure}[b]{0.49\textwidth}
        \centering
    	\includegraphics[width=\linewidth]{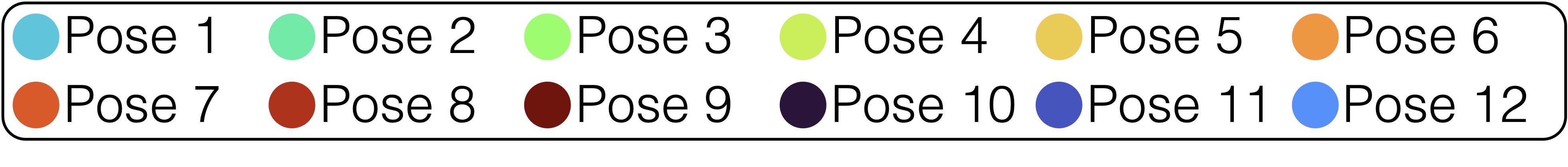}
    \end{subfigure}
    \par
    \begin{subfigure}[b]{0.49\textwidth}
        \centering
    	\includegraphics[width=\linewidth]{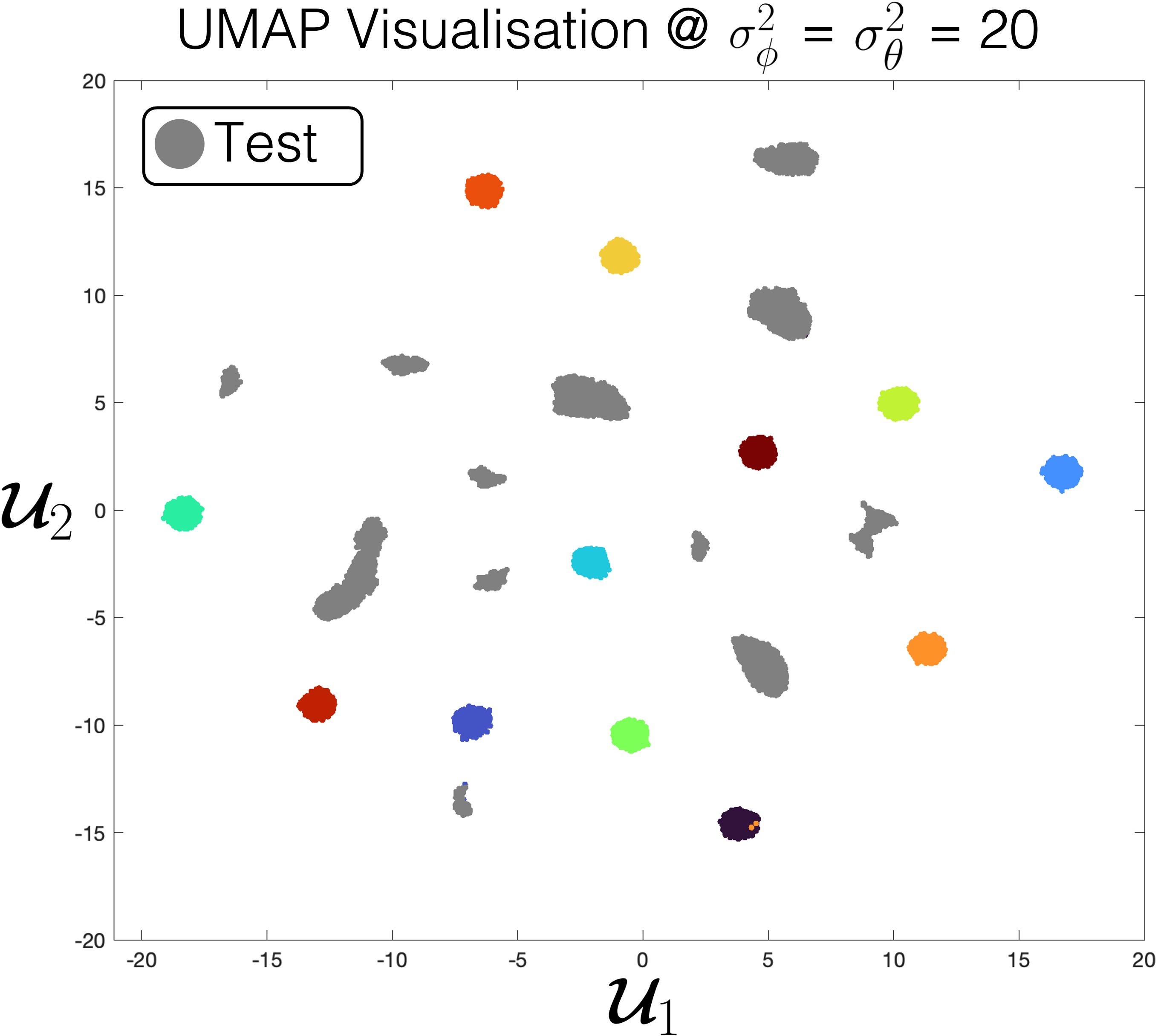}
    \end{subfigure}
    \begin{subfigure}[b]{0.49\textwidth}
        \centering
    	\includegraphics[width=\linewidth]{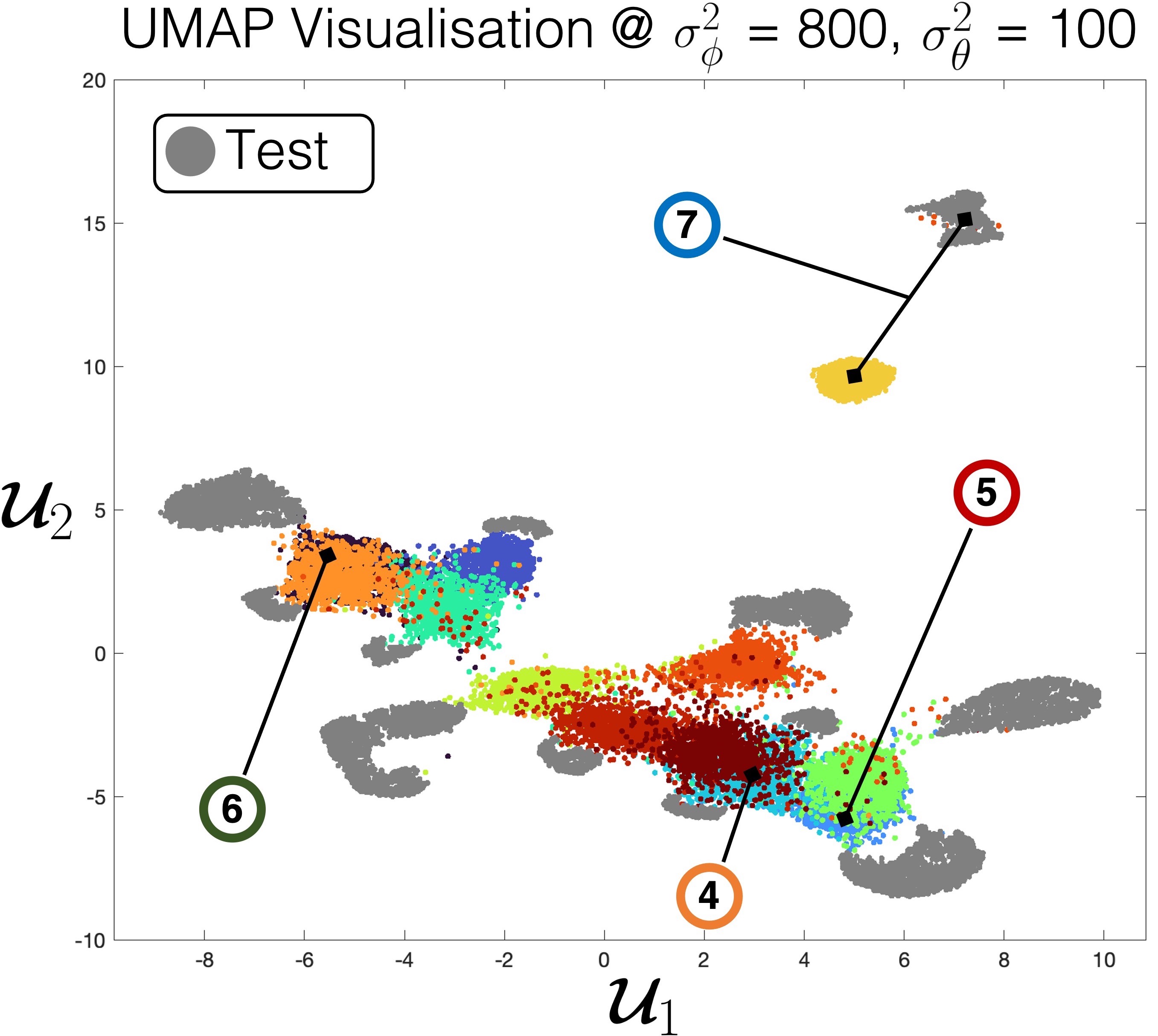}
    \end{subfigure}
    \par
    \begin{subfigure}[b]{0.49\textwidth}
        \centering
    	\includegraphics[width=\linewidth]{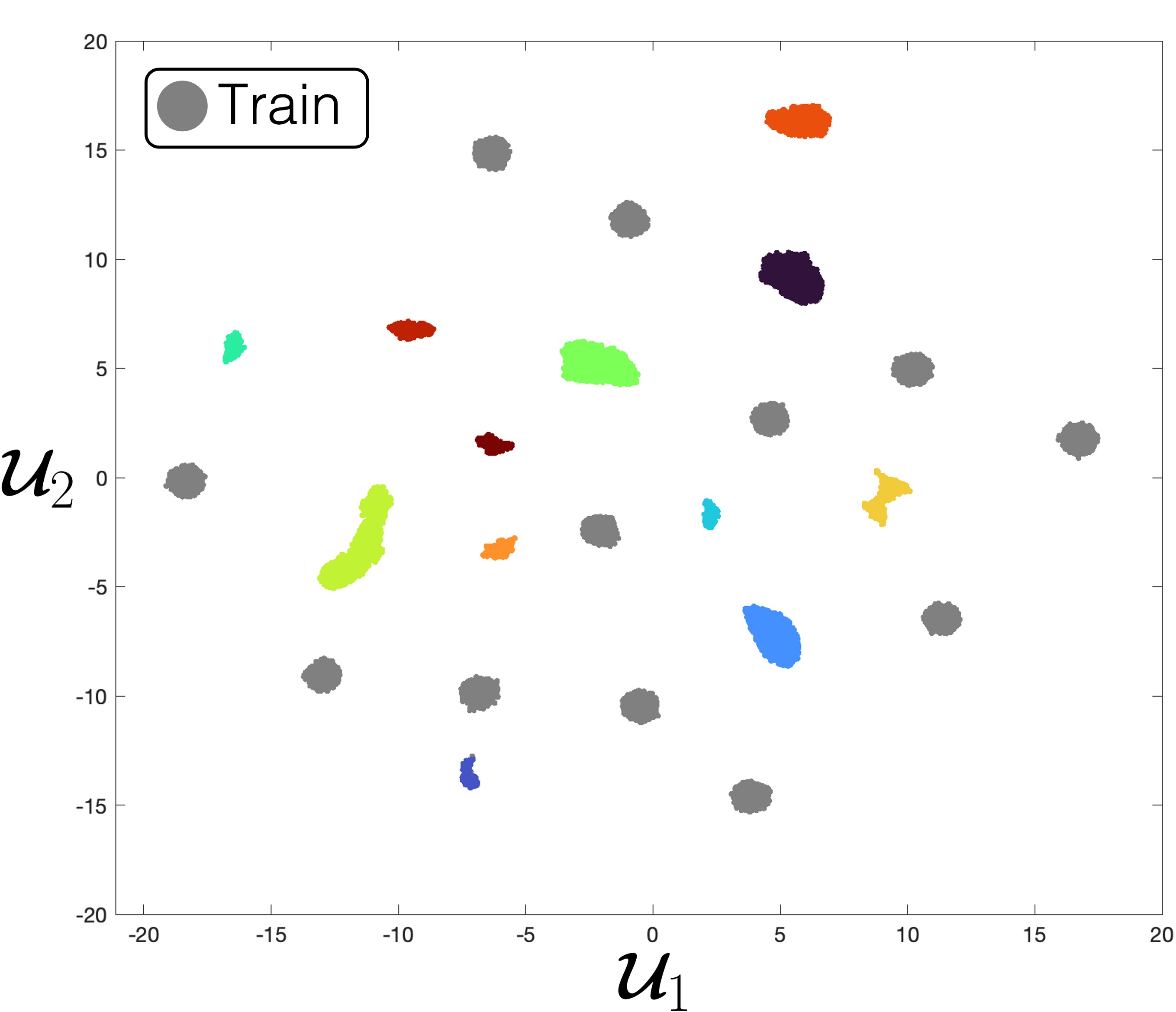}
    	\caption[Optional Caption]{}
    	\label{fig:UMAP2020}
    \end{subfigure}
    \begin{subfigure}[b]{0.49\textwidth}
        \centering
    	\includegraphics[width=\linewidth]{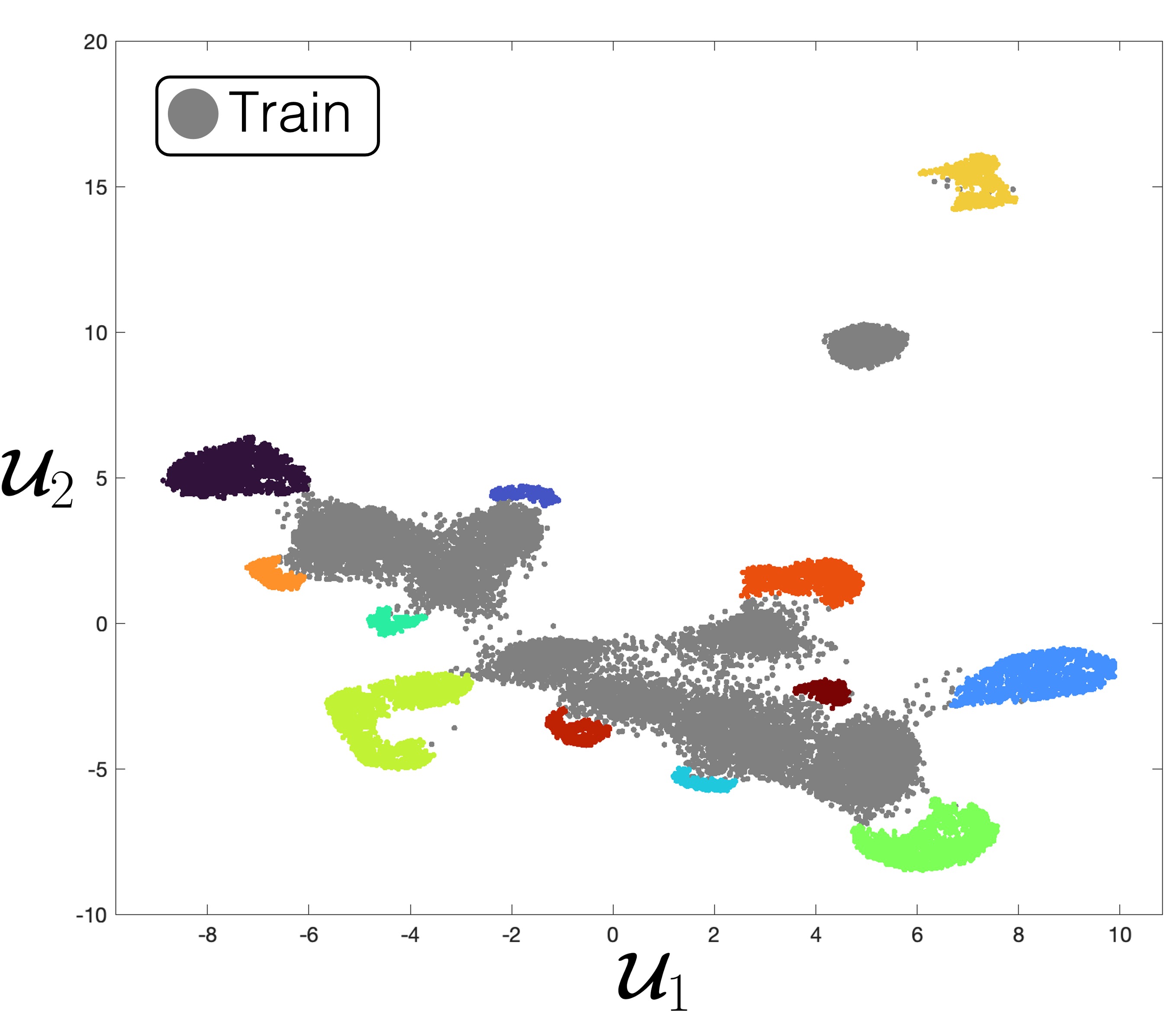}
    	\caption[Optional Caption]{}
    	\label{fig:UMAP800100}
    \end{subfigure}
    \caption{UMAP visualisations of $\prescript{_{\cross[0.4pt]}}{}{\bm \Psi}^w({\bm \tau})$ and $\prescript{*}{}{\bm \Psi}^w({\bm t})$ at (a) mild noise injection, and (b) optimal axis-dominated augmentation. For each row of figure, either train- or test-labelled datapoints are coloured based on their posture labels, while the others are greyed out.}
    \label{fig:UMAP}
\end{figure*}

\cref{fig:realmeanandstdaccuracyandF1} shows the results obtained via the proposed sleep posture learning framework.
Owing to the class-imbalanced $\prescript{*}{}{\bm \Psi}^w({\bm t})$, both $m_{acc}$ and $m_{F1}$ scores are reported. It is evident that the data augmentation settings have a substantial influence over the classification performance, with $m_{F1}$ ranging roughly between $60\%\ \text{and}\ 90\%$. Given mild noise injection at $(\sigma_\phi^2,\sigma_\theta^2) = (20,20)$, the mean $m_{F1}$ score was found to be $72.2\%$.


\begin{figure*}[!b]
    \centering
    \includegraphics[width=0.9\textwidth]{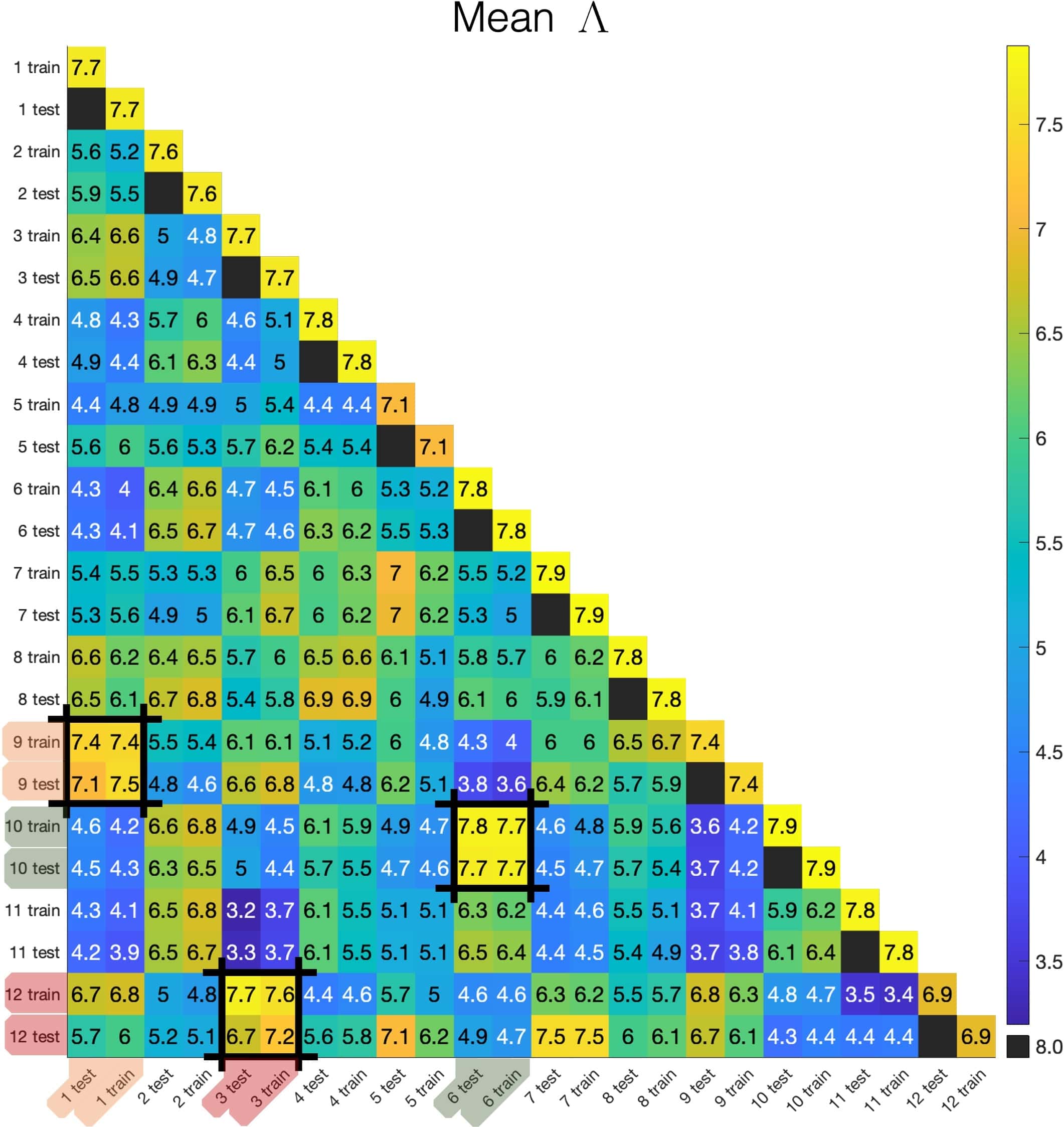}
    \caption{Intra-posture similarity matrix. Black squares highlight high similarity regions.}
    \label{fig:SimilarityMatrix}
\end{figure*}

Examining the $m_{F1}$ heat map of the human participant study provides a good picture of how augmentation hyperparameter tuning influences the classification performance.
In the virtual experiment, each classifier was trained and tested on augmented observations sharing the same postural variability. For the participant study, different data augmentation settings are used to pre-train multiple classifier models that are later tested on the same testing posture dataset, $\prescript{*}{}{\bm \Psi}^w({\bm t})$, with unknown train-test pose discrepancy. Logically speaking, a good data augmentation setting would then be one that produces an augmented training posture dataset of variability level close to the actual train-test pose discrepancy. Based on this, recommendations for optimal tuning of the data augmentation hyperparameters can be informed.


To facilitate understanding, let us subdivide ${\bm \Re}$ in \cref{fig:realmeanandstdaccuracyandF1} into three subgrids annotated by \circled{1}, \circled{2} and \circled{3} to study the effect of different augmentation settings on the classification performance. Subgrid \circled{1} shows that angle-dominant augmentation yields performance metrics similar to that of $(\sigma_\phi^2,\sigma_\theta^2) = (20,20)$. For subgrid \circled{2}, the performance undergoes a falling trend ($m_{F1}$ as low as $60\%$) in response to augmenting the axes and angles of rotation simultaneously. Finally, subgrid \circled{3} showcases further performance enhancement brought by axis-dominant augmentation, boosting $m_{F1}$ to about $90\%$. The reason why augmenting axes is more useful may be related to the presence of environmental objects (e.g. mattress and pillow) which constrain joint rotations, hence, variation in sleep postures rather owes to deviations in joint axes of rotation. Consequently, subgrid \circled{3} is recommended for data augmentation, specifically given $\sigma_\phi^2 \geqslant 800 \cap \sigma_\theta^2 = 100$.


To further understand how axis-dominant augmentation can contribute up to $30\%$ gain in performance compared to other augmentation settings, we use UMAP-empowered data visualisation described in \cref{sec:PerformanceInterpretation}. \cref{fig:UMAP} reports $\mathbfcal{U}$ for two different scenarios: mild noise injection at $(\sigma_\phi^2,\sigma_\theta^2) = (20,20)$ (\cref{fig:UMAP2020}), and optimal axis-dominated augmentation at $(\sigma_\phi^2,\sigma_\theta^2) = (800,100)$ (\cref{fig:UMAP800100}). At mild noise injection, \cref{fig:UMAP2020} shows large discrepancies between training and testing observations across most sleep postures, as reflected by the sparse distribution of scattered clusters of observations. This clarifies why, in the absence of a sufficiently large dataset, it is hard to accomplish satisfactory posture classification performance. On the other hand, \cref{fig:UMAP800100} showcases the effectiveness of the axis-dominant augmentation in bringing about structure to the data distribution as the training and testing observations of each posture are located in close proximity. The resultant classifier-friendly data distribution stands behind the significant rise in performance $(m_{acc}=92.7\%)$ and robustness to postural discrepancies compared to the mild augmentation case. Additionally, it is noteworthy how overlaps emerged between certain postures as in annotated regions \circled{4}, \circled{5} and \circled{6}.

\begin{figure*}[!b]
    \centering
    \begin{subfigure}[b]{0.6\textwidth}
        \centering
    	\includegraphics[width=\linewidth]{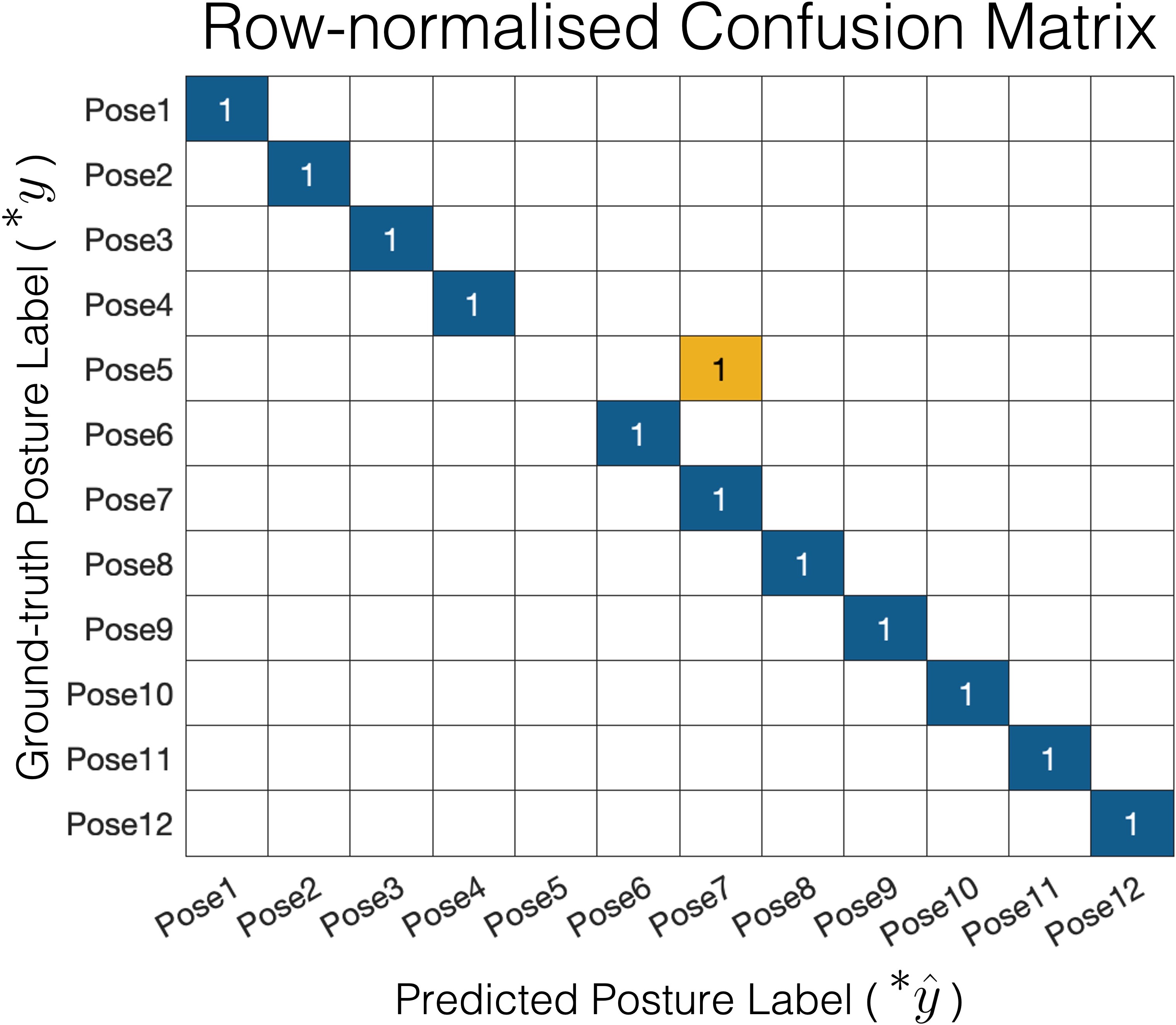}
    	\caption{}
    \label{fig:ConfusionMatrix800_100}
    \end{subfigure}
    \par
    \begin{subfigure}[b]{0.49\textwidth}
        \centering
    	\includegraphics[width=\linewidth]{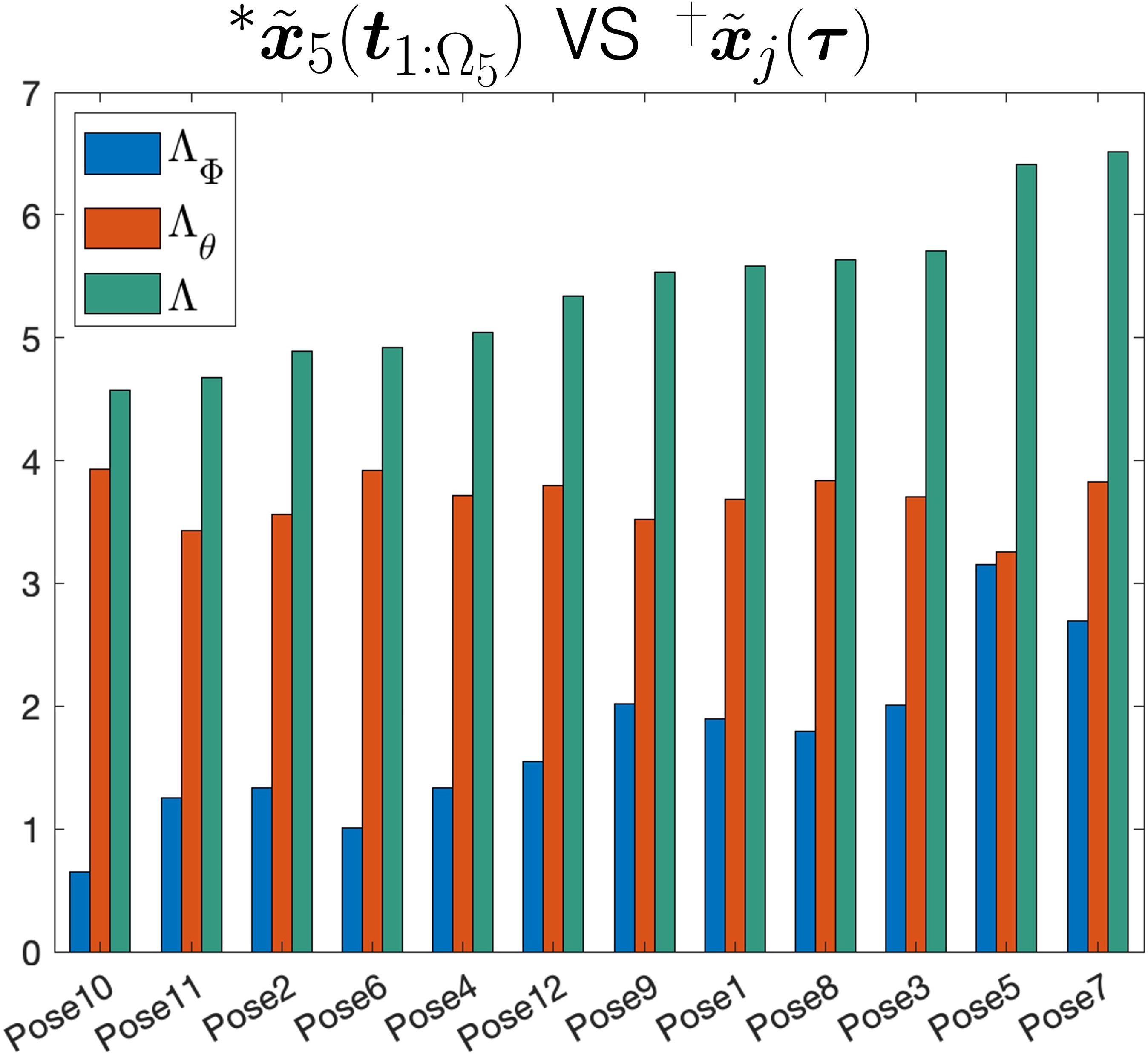}
    	\caption{}
    \label{fig:LambdaPose5Test}
    \end{subfigure}
    \begin{subfigure}[b]{0.49\textwidth}
        \centering
    	\includegraphics[width=\linewidth]{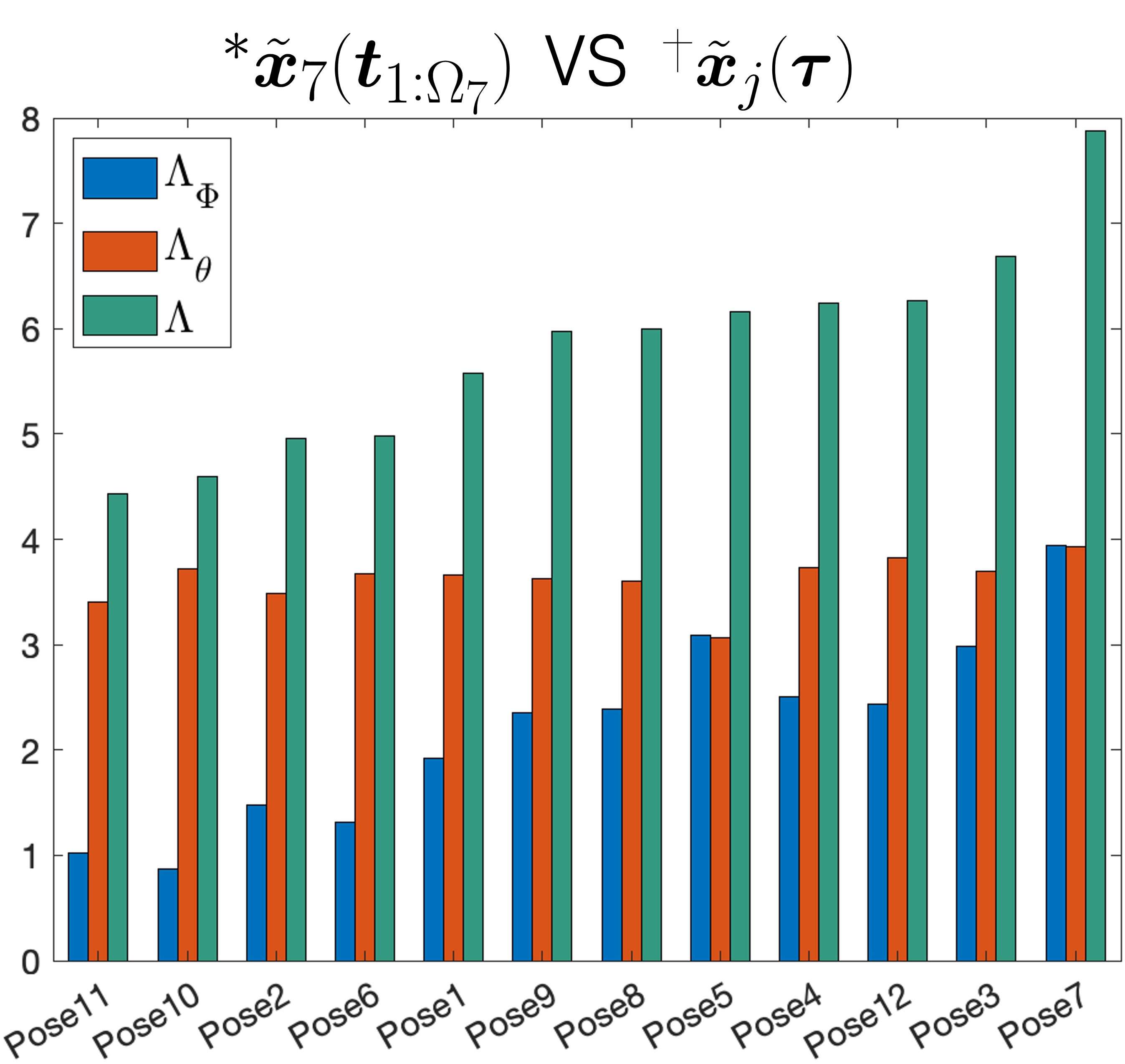}
    	\caption{}
    \label{fig:LambdaPose7Test}
    \end{subfigure}
    \caption{Performance evaluation and interpretation given optimal augmentation settings $(\sigma_\phi^2,\sigma_\theta^2) = (800,100)$: (a) row-normalised confusion matrix, and (b), (c) ``{\itshape one testing versus all training}'' analysis corresponding to $\mathbfcal{Y}_5$ and $\mathbfcal{Y}_7$.}
    \label{fig:EvalInterpretAug800_100}
\end{figure*}

\begin{figure*}[!b]
    \centering
    \includegraphics[width=\textwidth]{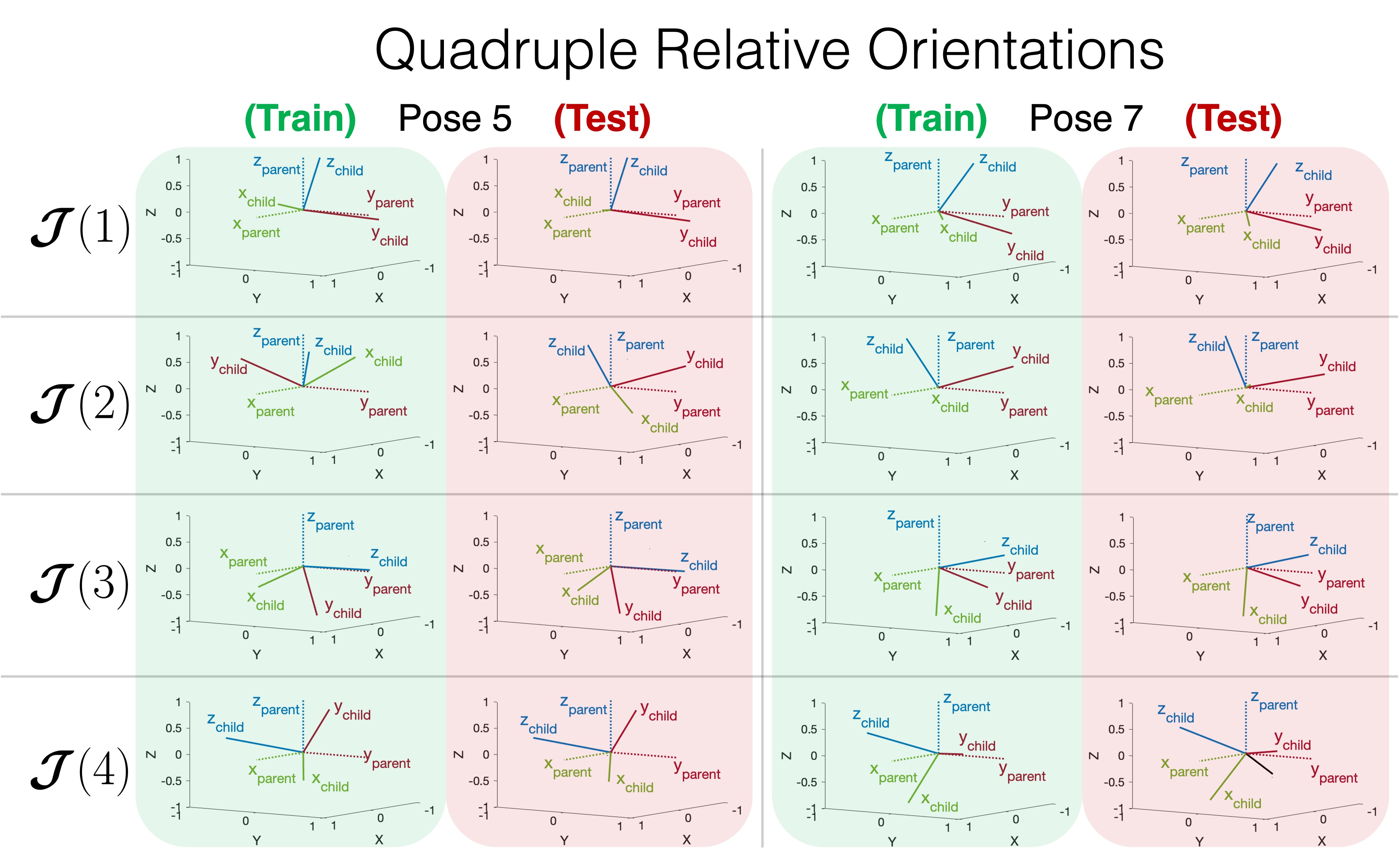}
    \caption{Mean joint orientations of training and testing observations for $\mathbfcal{Y}_5$ and $\mathbfcal{Y}_7$.}
    \label{fig:QuadrupleOrientationPose5vsPose7}
\end{figure*}

The hybrid metric $\Lambda$ proposed in \cref{sec:PerformanceInterpretation} can also be used to further understand the intra-posture similarities between: (i) $\mathbfcal{Y}_1\leftrightarrow\mathbfcal{Y}_9$, (ii) $\mathbfcal{Y}_3\leftrightarrow\mathbfcal{Y}_{12}$, and (iii) $\mathbfcal{Y}_6\leftrightarrow\mathbfcal{Y}_{10}$. \cref{fig:SimilarityMatrix} presents the mean $\Lambda$ with ${\bm x}_a$ and ${\bm x}_b$ exhausting all combinations of posture-specific (augmented) training and testing observations given the mild augmentation case $(\sigma_\phi^2,\sigma_\theta^2) = (20,20)$. Remarkably, the metric $\Lambda$ is capable of revealing postural similarities that UMAP did not capture in \cref{fig:UMAP2020}. Sifting data for such correlations and trends is of great significance to researchers and clinicians in terms of rethinking human postural analysis, for instance, to evaluate the efficacy of pose characterisation methods. Another possible usage of this map is the examination of posture definitions and confirming their parametric and subjective distinction from other postures before including it in the study.

\cref{fig:ConfusionMatrix800_100} shows the confusion matrix given the optimal augmentation setting; $(\sigma_\phi^2,\sigma_\theta^2) = (800,100)$. The SVM-ECOC model achieves 100\% classification accuracy on testing observations of all postures except $\mathbfcal{Y}_5$, demonstrating satisfactory robustness to the postural overlaps outlined in \cref{fig:UMAP800100,fig:SimilarityMatrix}. Such overlaps were viewed as a challenge in similar studies although these works considered no more than eight postures of standard-to-moderate complexity, see for example \cite{Fallmann2017}. To understand why the model confuses $\mathbfcal{Y}_5$ with $\mathbfcal{Y}_7$, we conduct a $\Lambda$-based similarity assessment specifically focused on the misclassification in \cref{fig:LambdaPose5Test}. Since the model relies on augmented training data to handle unseen testing observations, we compare ${\bm x}_a = \prescript{*}{}{\tilde{\bm x}}_5(\bm t_{1:\Omega_5})$ against all ${\bm x}_b = \prescript{_{\cross[0.4pt]}}{}{\tilde{\bm x}}_j(\bm \tau)\ \forall j \in \left[1,12\right]$, where $\tilde{(\cdot)}$ denotes the mean in time domain.
\cref{fig:LambdaPose5Test} reveals a small difference in $\Lambda$ between $\mathbfcal{Y}_5$ and $\mathbfcal{Y}_7$. Moreover, $\Lambda$ scores of both $\mathbfcal{Y}_5$ and $\mathbfcal{Y}_7$ indicate a moderate similarity level only around $6.5$ out of $8.0$, with no clear winner. Recalling region \circled{7} from \cref{fig:UMAP800100}, the relatively large train-to-test distance for $\mathbfcal{Y}_5$ confirms the participant was (unintentionally) inconsistent in replicating that posture during data collection, which again explains the misclassification of $\prescript{*}{}{\bm x}_5(\bm t)$. Further inspection into the root cause of such discrepancy is presented in \cref{fig:QuadrupleOrientationPose5vsPose7} which shows that the participant's $\mathbfcal{J}(2)$ mean orientation differed considerably between train- and test-labelled recordings of $\mathbfcal{Y}_5$. Such inexact posture recreation by participants is an occasional challenge inherent to similar works, as in \cite{Kwasnicki2018}.

For the sake of comparison, \cref{fig:LambdaPose7Test} shows the result of the same assessment of \cref{fig:LambdaPose5Test} but with ${\bm x}_a = \prescript{*}{}{\tilde{\bm x}}_7(\bm t_{1:\Omega_7})$. \cref{fig:LambdaPose7Test} reveals an uncertainty-free scenario where $\mathbfcal{Y}_7$ has a similarity metric of about $7.9$ out of $8.0$. Therefore, the $\Lambda$ metric can be regarded as a confidence measure associated with the output posture label, indicating how far one can trust the system at any instant of time.

Interestingly, \cref{fig:LambdaPose5Test,fig:LambdaPose7Test} shows that $\Lambda_\phi$ experience more acute variations in comparison to $\Lambda_\theta$. This clarifies why axis-dominant augmentation accomplishes better enhancement to the performance compared to angle-dominant augmentation. Such observation also reflects on the nature of in-bed postural analysis as environmental constraints essentially inhibit the mobility of joints, causing variation to mostly take place along the axial component.

For reference, posture classification using only the real training data available was conducted to directly evaluate the {\itshape simulation-to-real} (Sim2Real) gap. In this case, the training of the classification model is not limited to only one observation per posture (one shot). Instead, the SVM-ECOC model leverages the whole length of train-labelled posture recordings and utilises all observed segment-to-segment orientations for posture modelling. After 10 repeated train-test runs, the average classification accuracy of the model on $\prescript{*}{}{\bm \Psi}^w({\bm t})$ was found to be around $48.1\%$. The macro-averaged $m_{F1}$ is found to be $40.1\%$, which means the Sim2Real gap is $40\%$ to $60\%$. This chance-level accuracy reveals poor generalisation performance and justifies the need for a posture learning framework that better deals with data insufficiency - the aim of this paper. The proposed framework indeed provides a boost in $m_{F1}$ by more than 20\% out of the box and before fine-tuning the data augmentation hyperparameters.

\subsection{Comparison with state of the art}  \label{sec:standingandfuturedirections}
To the best knowledge of the authors, the presented work serves as the most advanced sleep posture analysis in the literature with twelve complex postures representing non-standard postural variations common during sleep. Similar works only covered no more than eight postures, many of which are minor variations of the four standard sleep postures. The protocol of sleep data collection is another crucial aspect that sets the presented methodology apart from others reported in the literature. Specifically, randomised strategies for both pose shuffling and train/test trial assignment are adopted to ensure statistical independence and to account for the participant gaining familiarity with the experiment over time. Moreover, the proposed one-shot posture learning framework makes the use of wearable technologies far easier and more viable, as it removes the need for expensive training data collection sessions. Lastly, the exclusive use of inertial sensor fusion brings forward approximate segment orientations instead of the rudimentary raw data approach often adopted in the literature. Therefore, our approach provides a more comprehensible posture representation to non-technical experts such as clinicians. To better highlight the benefits of the framework proposed in this paper, a comparison with existing works is summarised in \cref{tab:comparisonagainstexistingmethods}.

Some state-of-the-art studies \cite{Kwasnicki2018,Fallmann2017} had reported the possible occurrence of intra-posture similarity and inexact posture recreation, and regarded these as limitations without a clear attempt of formally verifying or quantifying them. A distinctive highlight of this work is the proposed interpretation approaches which provide qualitative and quantitative insights into the nature of sleep postures, their augmentation, and the classification problem as a whole. Our investigation shows that appropriate augmentation settings can make the classification robust to intra-posture similarity.

\newpage

\section{Conclusions}   \label{sec:conclusions}
A novel human sleep posture learning framework is proposed, capable of classifying twelve complex sleep postures. This goes beyond related works that mostly consider only four standard postures (supine, prone and lateral positions). The framework was first developed and tested through {\itshape in silico} sleep simulation, then successfully validated in a pilot human participant study. In both experimental pipelines, aggregate segment-to-segment orientations from four distal joints (wrists and ankles) were used to characterise the body posture. This simplified representation was the basis for the sleep posture learning task assigned to an ensemble classifier model. Computer graphics software and custom-made wearable sensor modules with inertial sensing capability were used, respectively, in the virtual and participant pipelines. A major highlight of this work is the use of inertial sensor fusion to gauge segment orientations instead of the raw sensor readings heavily used in the literature. Therefore, our posture representation is more comprehensible to non-technical end users, such as clinicians. Another prominent contribution of this work is the augmentation of postural observations which accelerated posture modelling with increased robustness given only one observation (shot) per posture, omitting the need for longitudinal data collection. The proposed one-shot learning scheme was found to boos the posture classification performance by up to $50\%$ with respect to learning from scarce postural observations. Despite insufficient training data and diversified posture selection, we report performance comparable to the state-of-the-art works. Lastly, we outlined a new metric-based approach and used it along with data visualisation to extract quantitative and qualitative insights on postural analysis, the added value of data augmentation, and the interpretation of the classification performance. The results carry evidence-backed findings that could potentially inform policies and recommendations for the use of wearable sensors in sleep medicine.

A number of directions may guide future research. Since inexact posture recreation (i.e. human non-compliance) seems to be a likely-to-happen open challenge, it might be useful to avoid the discretisation of the human posture space in classification problems and resort to partial- or full-body posture estimation instead. A potentially interesting work would be to examine the system performance when further increasing the number and complexity of sleep postures, and checking whether the robustness to intra-posture similarity will continue to hold. Another future direction could investigate whether incorporating additional absolute segment orientations at the pose characterisation stage would strengthen the differences between postures and reduce overlap.

\section*{Acknowledgements}   \label{sec:acknowledgements}
The authors would like to thank Daniel Potts for his assistance with the development of the wearable sensors.

\section*{Funding}   \label{sec:funding}
Omar Elnaggar (first author) is supported by the  University of Liverpool  Doctoral Network in AI for Future Digital Health.

\section*{Declaration of Competing Interest}   \label{sec:CompetingInterest}
All authors declare that they have no conflict of interest.

\bibliography{IFrefs11}

\begin{table*}[b]
    
    \centering
    \caption{Comparison with existing works in the literature.}
    \label{tab:comparisonagainstexistingmethods}
    \footnotesize
    \def\arraystretch{1.3}%
    \begin{tabularx}{1.00\textwidth}{@{}*{9}{L}@{}}
    
    \toprule
    Reference & \multicolumn{8}{c}{Comparison Criteria} \\
    \cmidrule{2-9}
    & Sleep \newline Postures & Classification \newline Algorithm & Sensor \newline Placement & Dataset \newline Duration & Data \newline Augmentation & Sensor Fusion & ${m_{acc}}$ \newline ${m_{F1}}$ & Additional \newline Information \\
    \midrule
    Zhang et al. \newline \cite{Zhang2015} 
    & 4 \newline (standard) 
    & LDA   
    & 1 IMU \newline (chest)  
    & - 
    & No  
    & No    
    & 99\% \newline - 
    & Heart rate; respiratory rate \\ 
    
    Sun et al. \newline \cite{Sun2017} 
    & 4 \newline (standard) 
    & na\"ive Bayes, Bayesian network, DT, Random Forest   
    & 1 IMU \newline (left wrist)  
    & 70 nights 
    & No  
    & No    
    & (60.3-91.8)\% \newline - 
    & Respiratory rate \\ 
    
    Eyobu et al. \newline \cite{Eyobu2018a} 
    & 4 \newline (standard) 
    & LSTM   
    & 1 IMU \newline (upper arm)  
    & - 
    & No  
    & No    
    & 99\% \newline - 
    & - \\ 
    
    Alinia et al. \newline \cite{Alinia2020} 
    & 4 \newline (standard) 
    & Adaptive LSTM   
    & 1 IMU \newline (9 locations)  
    & $>$56 min. 
    & No  
    & No    
    & (64.9-98.4)\% \newline (62.9-98.2)\% 
    & - \\ 
    
    Alinia et al. \newline \cite{Alinia2020} 
    & 4 \newline (standard) 
    & Ensemble tree classifier   
    & 1 IMU \newline (9 locations)  
    & $>$56 min. 
    & No  
    & No    
    & (62.9-94.4)\% \newline (60.9-93.6)\% 
    & - \\ 
    
    Jeon et al. \newline \cite{Jeon2019} 
    & 4 \newline (standard) 
    & Dynamic state transition framework   
    & 3 IMUs \newline (chest and wrists)  
    & - 
    & No  
    & No    
    & 94\% \newline 79\% 
    & In-bed motion recognition \\ 
    
    Monroy et al. \newline \cite{BernalMonroy2020} 
    & 4 major; \newline 2 minor \newline (standard) 
    & $k$-NN, SVM   
    & 3 IMUs \newline (chest and ankles)  
    & $\sim$60 min. 
    & No  
    & No    
    & - \newline 100\% 
    & Alert need for postural change \\ 
    
    Monroy et al. \newline \cite{BernalMonroy2020} 
    & 4 major; \newline 2 minor \newline (standard) 
    & DT   
    & 3 IMUs \newline (chest and ankles)  
    & $\sim$60 min. 
    & No  
    & No    
    & - \newline 51\% 
    & Alert need for postural change \\ 
    
    Kwasnicki et al. \newline \cite{Kwasnicki2018} 
    & 4 major; \newline 4 minor \newline (moderate) 
    & LDA, $k$-NN, na\"ive Bayes, DT   
    & 3 IMUs \newline (chest and wrists)  
    & $\sim$160 sec. 
    & No  
    & No    
    & 92.5\% \newline - 
    & Sleep phase prediction \\ 
    
    Fallmann et al. \newline \cite{Fallmann2017} 
    & 4 major; \newline 4 minor \newline (moderate) 
    & GMLVQ   
    & 3 IMUs \newline (chest and ankles)  
    & $>$3.75 hr. 
    & No  
    & No    
    & (78-99.8)\% \newline - 
    & Evaluation at different settings \\ 
    
    Fallmann et al. \newline \cite{Fallmann2017} 
    & 4 major; \newline 2 minor \newline (standard) 
    & GMLVQ   
    & 3 IMUs \newline (chest and ankles)  
    & $\sim$7 hr. 
    & No  
    & No    
    & (58-98)\% \newline - 
    & Evaluation at different settings \\ 
    
    Chang et al. \newline \cite{Chang2018} 
    & 4 \newline (standard) 
    & $k$-NN   
    & 1 IMU \newline (left wrist)  
    & $\sim$6 hr. 
    & No  
    & No    
    & $>$90\% \newline - 
    & Nocturnal behavioural analysis \\ 
    
    Our Approach \newline (Virtual sleep) 
    & 12 \newline (non-standard, complex) 
    & SVM-ECOC   
    & N/A  
    & N/A   
    & Yes  
    & Yes    
    & (81-100)\% \newline (81-100)\% 
    & - \\ 
    
    Our Approach \newline (Participant study w/o one-shot learning) 
    & 12 \newline (non-standard, complex) 
    & SVM-ECOC   
    & 4 dual-IMU modules \newline (wrists and ankles)  
    & $\sim$24 min.
    & No  
    & Yes    
    & 48.4\% \newline 40.1\% 
    & - \\ 
    
    Our Approach \newline (Participant study with one-shot learning) 
    & 12 \newline (non-standard, complex) 
    & SVM-ECOC   
    & 4 dual-IMU modules \newline (wrists and ankles)  
    & $\nicefrac{1}{30}$ sec. (one shot per pose for training) Total dataset($\sim$24 min.)
    & Yes  
    & Yes    
    & (73.9-92.7)\% \newline (72.2-89.7)\% 
    & Postural similarity; visualisation of posture datasets \\ 
    \bottomrule
    
    \end{tabularx}
    
\end{table*}

\end{document}